\documentclass[pdflatex,sn-apa]{sn-jnl}

\usepackage{graphicx}%
\usepackage{multirow}%
\usepackage{amsmath,amssymb,amsfonts}%
\usepackage{amsthm}%
\usepackage{mathrsfs}%
\usepackage[title]{appendix}%
\usepackage[pdftex,dvipsnames]{xcolor}%
\usepackage{textcomp}%
\usepackage{manyfoot}%
\usepackage{booktabs}%
\usepackage{algorithm}%
\usepackage{algorithmicx}%
\usepackage{algpseudocode}%
\usepackage{listings}%
\usepackage{mathtools}

\usepackage{comment}
\usepackage{tabularx}
\usepackage{hyperref}
\usepackage{microtype}
\usepackage{multirow}
\usepackage{caption}
\usepackage{array}
\usepackage{hhline}
\usepackage{caption}
\usepackage{subcaption}
\usepackage{lipsum}
\usepackage{xcolor}
\usepackage{placeins}

\usepackage[inline]{enumitem}

\usepackage{xargs} 
\usepackage[colorinlistoftodos,prependcaption,textsize=tiny]{todonotes}
\newcommandx{\unsure}[2][1=]{\todo[linecolor=red,backgroundcolor=red!25,bordercolor=red,#1]{#2}}
\newcommandx{\change}[2][1=]{\todo[linecolor=blue,backgroundcolor=blue!25,bordercolor=blue,#1]{#2}}
\newcommandx{\info}[2][1=]{\todo[linecolor=OliveGreen,backgroundcolor=OliveGreen!25,bordercolor=OliveGreen,#1]{#2}}
\newcommandx{\improvement}[2][1=]{\todo[linecolor=Plum,backgroundcolor=Plum!25,bordercolor=Plum,#1]{#2}}
\newcommandx{\thiswillnotshow}[2][1=]{\todo[disable,#1]{#2}}

\DeclareMathOperator*{\argmax}{arg\,max}

\theoremstyle{thmstyleone}%
\theoremstyle{thmstyletwo}%

\theoremstyle{thmstylethree}%

\begin{document}

\title[EMTeC: A Corpus of Eye Movements on Machine-Generated Texts]{EMTeC: A Corpus of Eye Movements on Machine-Generated Texts}


\author*[1]{\fnm{Lena S.} \sur{Bolliger}}\email{bolliger@cl.uzh.ch}

\author[1]{\fnm{Patrick} \sur{Haller}}\email{haller@cl.uzh.ch}

\author[1]{\fnm{Isabelle C. R.} \sur{Cretton}}\email{isabellecarolinerose.cretton@uzh.ch}

\author[1,2]{\fnm{David R.} \sur{Reich}}\email{david.reich@uni-potsdam.de}

\author[1]{\fnm{Tannon} \sur{Kew}}\email{kew@cl.uzh.ch}

\author[1,2]{\fnm{Lena A.} \sur{Jäger}}\email{jaeger@cl.uzh.ch}

\affil*[1]{\orgdiv{Department of Computational Linguistics}, \orgname{University of Zurich}, \orgaddress{\street{Andreasstrasse 15}, \city{Zurich}, \postcode{8050}, \country{Switzerland}}}

\affil[2]{\orgdiv{Department of Computer Science}, \orgname{University of Potsdam}, \orgaddress{\street{An der Bahn 2}, \city{Potsdam}, \postcode{14476}, \country{Germany}}}


\abstract{

The \textbf{E}ye Movements on \textbf{M}achine-Generated \textbf{Te}xts \textbf{C}orpus (EMTeC) is a naturalistic eye-movements-while-reading corpus of 107 native English speakers reading machine-generated texts. The texts are generated by three large language models using five different decoding strategies, and they fall into six different text type categories. EMTeC entails the eye movement data at all stages of pre-processing, \emph{i.e.}, the raw coordinate data sampled at 2000\,Hz, the fixation sequences, and the reading measures. It further provides both the original and a corrected version of the fixation sequences, accounting for vertical calibration drift. Moreover, the corpus includes the language models' internals that underlie the generation of the stimulus texts: the transition scores, the attention scores, and the hidden states. The stimuli are annotated for a range of linguistic features both at text and at word level. We anticipate EMTeC to be utilized for a variety of use cases such as, but not restricted to, the investigation of reading behavior on machine-generated text and the impact of different decoding strategies; reading behavior on different text types; the development of new pre-processing, data filtering, and drift correction algorithms; the cognitive interpretability and enhancement of language models; and the assessment of the predictive power of surprisal and entropy for human reading times. The data at all stages of pre-processing, the model internals, and the code to reproduce the stimulus generation, data pre-processing and analyses can be accessed via \url{https://github.com/DiLi-Lab/EMTeC/}.

}




\maketitle




\section{Introduction}
\label{sec:introduction}


Human eye movements in reading provide insight into the cognitive mechanisms involved in human language processing~\citep{rayner1998} and reveal information about key properties and structures of the text being read~\citep{rayner2009, engbert2005swift, reichle1998toward}. As a consequence, the investigation and application of readers' eye movements has experienced a great upswing within the past two decades across a variety of fields, including experimental and computational psycholinguistics, cognitive psychology, education science, natural language processing (NLP),  and various areas of computer science, including human-computer interaction, robotics, and artificial intelligence. 

\paragraph{Psycholinguistic Analysis of Eye Movements in Reading}

Within the field of psycholinguistics, most eye-tracking studies aim at investigating very clearly-defined and usually theoretically motivated hypotheses about human reading and language comprehension processes. These hypotheses are frequently studied by using minimal pair stimuli that allow for a targeted manipulation of the linguistic construction under investigation, thereby enabling to disentangle specific phenomena and isolate the ones that, based on clear theoretical assumptions, provide the answers to the hypotheses. This is an important and effective approach to studying linguistic constructions and their effect on reading behavior and the cognitive processes involved, especially as theoretically relevant constructions that allow for teasing apart competing theories are often complex and infrequently occurring in natural language.

However, research has also underlined the importance of studying language processing in a naturalistic setting~\citep{demberg2008data, demberg2019cognitive, nastase2020keep}: this entails moving away from minimal pair stimuli to ones that naturally occur in the language, such as newspaper articles or fictional texts. This is also accompanied by stimuli that extend beyond the sentence level and results in the possibility of examining reading behavior not only within a sentence but across sentence boundaries and entire paragraphs. 
While minimal pair stimuli usually serve the investigation of lexical properties (e.g., word length, lexical frequency, or word predictability), or of syntactic and semantic processing at the sentence level (e.g., garden-path effects, similarity-based interference in syntactic dependency formation, grammatical illusions, or effects of local coherence), naturalistic stimuli at the paragraph level allow for studying different linguistic phenomena and discourse processing beyond sentence boundaries. These include research on co-reference resolution beyond the sentence boundary~\citep{luo-glass-2018-learning}, the dependence of eye movement patterns on global text difficulty~\citep{rayner2006reflections}, and the inference of passage-level text comprehension in adults~\cite{reich2022inferring, prasse2024improving} and children~\citep{joseph2021inference}.
It is this expanding of controlled experiments with highly complex and constructed stimuli into the inclusion of naturalistic texts that has resulted in the usage of eye-tracking-while-reading data beyond the realm of psycholinguistics.

Another line of research that has been steadily growing 
involves surprisal theory~\citep{hale2001probabilistic, levy2008expectation}. 
The notion of surprisal operationalizes the relationship between language and cognitive effort. More specifically, surprisal quantifies the predictability of a word given its context.  Surprisal is defined as the negative log-probability of a word conditioned on its preceding linguistic and extra-linguistic context, and it has been argued to be proportional to the cognitive effort associated with the  processing of this word, typically measured in terms of reading times. This correlation of surprisal with behavioral  measurements in reading has been corroborated extensively~\citep[][\emph{inter alia}]{demberg2008data, shain2021cdrnn, hoover2023plausibility, pimentel2023effect}.


\paragraph{Leveraging Eye Movements in Reading for Technical Purposes}

In recent years, eye movements in reading have also been increasingly leveraged for more technical and application-oriented purposes and investigations.
First, the accompaniment of the naturalistic stimuli by other (quasi-)experimental variables that are not tied to the stimuli themselves but to the readers enables both the statistical investigation as well as the inference of reader-specific properties from the eye movement data, such as the detection of dyslexia~\citep{raatikainen2021detection, haller2022eye} or the prediction of reading comprehension~\citep{reich2022inferring, meziere2023using, prasse2024improving} and readers' language proficiency~\citep{berzak2018assessing}.

Second, eye-tracking-while-reading data has also been increasingly utilized within the field of Natural Language Processing (NLP) for various use cases. On the one hand, it has been utilized to enhance the performance of language models (LMs) on a variety of downstream tasks~\citep{hollenstein2020towards, deng2023-augmented, deng2024finetuning}, such as sentiment analysis~\citep{mishra2017leveraging, long2017cognition, yang2023plm, tiwari2023predict}; named entity recognition (NER)~\citep{hollenstein2019entity}; sentence compression and paraphrase generation~\citep{sood2020improving}; the generation of image captions~\citep{takmaz2020generating}; reading task classification~\citep{hollenstein2021reading}; co-reference resolution~\citep{cheri2016leveraging}; natural language inference, word sense disambiguation and question answering~\citep{wang2024gaze}; and improving machine comprehension~\citep{malmaud2020bridging}. 
On the other hand, eye movement data has also been employed to explore the cognitive plausibility of language models and the extent of their cognitive interpretability~\citep{keller2010cognitively, beinborn2023cognitive}. This line of research comprises the interpretation of neural attention and its alignment with human attention, quantified in terms of reading times~\citep{sood2020interpreting, bensemann2022eye, eberle2022transformer}; the evaluation of language models' abilities to predict human reading behavior~\citep{hollenstein2021multilingual, hollenstein2022patterns, merkx2020human}; and the assessment of the alignment of importance attributed to words in a sentence between models and humans~\citep{hollenstein2021relative}.

Third, with the advent of neural language models, the investigation of surprisal theory has also gained new momentum. Eye-tracking-while-reading data is now not only used to investigate the predictability effects as postulated by surprisal theory, but they are also employed as indicators of how well surprisal can be estimated in the first place. Since surprisal presupposes access to the probabilities of words given their context and thus to the true probability distribution over the vocabulary, which is unknown in practice, it can only ever be approximated. Earlier approaches in this area aimed to measure surprisal using \emph{n}-gram models~\citep{jurafskymartin2000, mitchell2010syntactic, fossum2012sequential} or Hierarchical Hidden Markov models~\citep{wu2010complexity} to approximate surprisal. Others have employed probabilistic context-free grammars (PCFGs), which are phrase structure grammars augmented with probabilities on the expansion rules~\citep{hale2003information, hale2006uncertainty, roark2009deriving}. Surprisal can also be empirically measured by means of Cloze tasks~\citep{taylor1953cloze, smith2013predictability} or sentence completion tasks~\citep{Jaeger_JML2015}. These methods are based on experiments with human participants during which they are given a context and have to guess the next or missing word. The probabilities are estimated as the proportion of subjects who correctly guess the word based on the context. More recent research has turned to the utilization of language models for the estimation of surprisal, such as Long Short-term Memory neural networks~\citep{goodkind2018predictive}. 
Since the advent of the Transformer architecture~\citep{vaswani2017attention}, decoder-based autoregressive language models such as GPT-2~\citep{radford2019language} have been a popular choice. 
This has initiated a parallel line of research into surprisal with a focus on which language models have the highest predictive power, \emph{i.e.}, which language model estimates surprisal in a way such that it has the best fit on human reading times. Studies have shown that there is a relationship between the predictive power of a language model and both its size in terms of parameters~\citep{oh2023does} and the size (and compositionality) of the data it has been trained on~\citep{oh-schuler-2023-transformer}. 
Additionally, \citet{haller2024language} also show that the predictive power differs with respect to the cognitive group for which the reading times are predicted.

And last, there has been an upsurge of machine learning-based generative models of human eye movements during recent years. While cognitive models of eye movements implement theories of reading, such as the E-Z Reader model~\citep{reichle1998toward}, the {\"U}ber-Reader model~\citep{veldre2020towards}, or the SWIFT model~\citep{engbert2005swift}, these more recent models are purely data-driven. They simulate human eye movements on new texts that are similar to the ones they have been trained on, with some predicting fixation sequences~\citep{nilsson2009learning, nilsson2011entropy, hahn2016modeling, wang2019new, Deng_Eyettention2023, bolliger-etal-2023-scandl} and others raw gaze  coordinates~\citep{Prasse_SP-EyeGAN2023,prasse2024improving}. For training such models, it is crucial to have a wide range of eye movement corpora available that span both different languages as well as different text types and linguistic characteristics.


\paragraph{Reading as a Cultural Skill: The Focus on \emph{What} is Being Read}

Eye movements in reading leveraged to investigate human language processing require --- given the very nature of the research hypotheses --- to place the focus on the experimental stimuli themselves, as many psycholinguistic premises can only be answered by manipulating said stimuli. However, the question of \emph{what} is being read can also be approached in more general terms. Reading is both a cultural skill as well as a socio-cognitive process; it is not only an intellectual activity that contributes to problem solving skills and the evolution of independent and critical thought processes~\citep{st2008reading}, but it is also the means by which people learn culturally appropriate ways of engaging with the written texts and consume culturally relevant and appropriate information and values~\citep{bloome1985reading}. As such, what is being read -- both the contents and their origins -- is of importance on a level beyond the individual reader. The advent of large language models (LLMs) such as Gemini~\citep{team2023gemini}, Llama 2~\citep{touvron2023llama2}, and GPT-4~\citep{achiam2023gpt} has enabled the generation of texts that are not only grammatically correct but also highly persuasive and nearly indistinguishable from human-written material~\citep{kumarage2024survey}. Although the utility of these models cannot be denied, having been adopted in various domains such as journalism and academia, they do not come without ethical challenges. The combination of generating coherent text and the absence of consciousness and a real author persona can erode public trust and distort societal perceptions~\citep{chakraborty2023possibilities}; the true origin of such generated texts remains unknown to the readers. This is exacerbated by the intuition that humans spend their lives developing a bias towards assuming that well-written and eloquent texts are trustworthy. LLMs reduce this to absurdity, as they bring the form to perfection while the content is merely a by-product of their training. The European Union has recently adopted the Artificial Intelligence Act\footnote{\url{https://artificialintelligenceact.eu/}}, according to which the labelling of AI-generated content is mandatory, a law which is overall very difficult to enforce. 
With reading being such a central component to both how values and information are distributed within a society as well as how said society is perceived by readers, it is imperative to investigate more closely how readers engage with texts generated by language models. As such, it might not only be relevant to inspect the output of these models, \emph{i.e.}, the texts they generate, and how people read those, but it might also be insightful to research \emph{how} the models generate those texts and how different generation strategies affect reading behavior. This would entail looking into how different decoding strategies, \emph{i.e.}, different ways of modulating the probability distributions over next words to be generated, affect the output texts and the way readers engage with them.

\paragraph{Introducing EMTeC}

In this paper, we present EMTeC, the \textbf{E}ye Movements on \textbf{M}achine-Generated \textbf{Te}xts \textbf{C}orpus, a naturalistic corpus of native English speakers' eye movements in reading machine-generated texts of different genres. EMTeC is a clear departure from previous eye movements in reading corpora, as the experimental stimuli were generated with three different large language models (LLMs) of different size and from different model families, and that each LLM generated texts belonging to six different text types using five different neural decoding strategies. 
Moreover, we not only release the eye movement data and the experimental stimuli, but we additionally provide the LLMs' internals: the transition scores, attention scores, and hidden states for each generated text. We prompted the models to generate a series of texts of 6 different text types (e.g., fiction, poetry, summarization, \dots ). For the resulting stimulus corpus we created comprehensive text- and word-level annotations, such as surprisal estimates, text difficulty metrics, part-of-speech tags, dependency tags, and more.  
Moreover, each stimulus is accompanied by a comprehension question and two rating questions, one on subjective text difficulty and one on engagement with the text, which are answered by the participants.

In order to promote transparency, reproducibility, and reusability of EMTeC, we not only release the eye-movement data at its final processed stage but also all intermediate steps of pre-processing. More specifically, this comprises the raw data consisting of coordinate-timestamp samples, the fixation sequence data (scanpaths) extracted from the raw data by means of a gaze-event detection algorithm, and the word-level reading measures data computed from the fixation sequences. Moreover, we provide both an uncorrected and a corrected version of the data. 
In the latter, possible vertical drifts due to calibration degradation in the fixation data has been manually corrected by reassigning fixations that have been mapped to the wrong area of interest (the wrong word) to the correct one.   
We make all code available in a reproducible format, which includes the implementations of the stimulus generation, the text- and word-level annotation, the drift correction pipeline, and all stages of eye-tracking data pre-processing.

In summary, EMTeC offers these main features:
\begin{itemize}
    \item The texts are machine-generated by LLMs of different sizes and different families using five of the most common decoding strategies;
    \item The texts belong to six different text types (non-fiction (argument and description), fiction (story and dialog), poetry, summarization, news article, key-word text)
    \item We release all LLM internals, \emph{i.e.}, the transition scores, attention scores, and hidden states;
    \item We release the eye movement data in all stages of pre-processing, \emph{i.e.}, raw coordinate data, fixation sequences, and reading measures;
    \item We release both the corrected and uncorrected eye movement data;
    \item We release text- and word-level annotations (surprisal, PoS tags, dependency tags, readability scores, ...)
    
    \item All code is made publicly available, allowing for transparency over the entire pipeline from stimulus generation to pre- and post-processing of the eye movement data.
\end{itemize}

\paragraph{On the Use Cases of EMTeC}

Below we present a (non-exhaustive) list of the use cases of EMTeC. There are different axes to the EMTeC data that can all be leveraged for a variety of intents and purposes: the eye movement data at different stages of pre-processing; the uncorrected and corrected data versions; the different text types; and the machine-generated nature of the texts and the model internals.

The \textbf{eye movement data at different pre-processing stages} includes the raw data consisting of coordinate-timestamp samples, the fixation sequence data, and the reading measures. As suggested by \citet{Jakobi-dataquality-2024}, we share the data at all three stages of pre-processing to allow for a broad range of use cases. \\

\noindent The \textbf{raw data} can be used for 
\begin{itemize}
    \item the development of pre-processing and data filtering methods such as blink detection algorithms~\citep{krolak2012eye, MORRIS2002129}; 
    \item the development of new gaze event detection algorithms, such as saccade and fixation detection algorithms~\citep{sauter1991analysis, salvucci2000identifying, smeets2003nature, olsson2007real, nystrom2010adaptive};
    \item the training and evaluation of generative models of human eye movements that generate raw data samples~\citep{Prasse_SP-EyeGAN2023, prasse2024improving};
    \item and research on vision and oculomotor control, such as the investigation of oculomotor micro-movements~\citep{engbert2006microsaccades, martinez2006microsaccades, martinez2009microsaccades} or the velocity and acceleration profiles of saccades~\citep{JANTTI20111476, BACHURINA2022e08826}.
\end{itemize}
Raw data provides researchers with the highest flexibility when using the data, allowing them for applying pre-processing algorithms different from the ones used in this paper. \\

\noindent The \textbf{fixation sequence data} can be used for
\begin{itemize}
    \item the training and evaluation of generative models of eye movements in reading that simulate scanpaths on textual stimuli~\citep{nilsson2009learning, nilsson2011entropy, hahn2016modeling, wang2019new, Deng_Eyettention2023, bolliger-etal-2023-scandl};
    \item the psycholinguistic analysis of scanpaths to gain insights into the cognitive processes involved in reading~\citep{VONDERMALSBURG2011109, von2012scanpaths, malsburg2013scanpaths};
    \item the enhancement of language models with scanpaths (gaze-augmented language models)~\citep{deng2023-augmented, khurana2023synthesizing, yang2023plm, deng2024finetuning};
    \item and the development of metrics that quantify the differences and similarities between scanpaths~\citep{cristino2010scanmatch, shepherd2010human, mathot2012simple, jarodzka2010vector, VONDERMALSBURG2011109}.
\end{itemize}

\noindent The \textbf{reading measures} together with the linguistic annotations of the stimulus texts can be used for 
\begin{itemize}
    \item statistical analyses to evaluate and develop psycholinguistic theories of eye movements in reading. This includes, but is not limited to, investigations of phenomena at the lexical level such as word-length and lexical frequency effects, or the syntactic level, such as surprisal effects~\citep{demberg2008data, shain2021cdrnn, hoover2023plausibility, pimentel2023effect}, and locality or anti-locality effects~\citep{demberg2008data, bartek2011search};\footnote{Eye-tracking corpora with naturalistic stimuli can be leveraged for researching a variety of psycholinguistic phenomena; they allow for evaluating theories without first explicitly designing an experiment.} 
    \item investigations of the predictive power of surprisal~\citep{oh2023does, oh-schuler-2023-transformer};
    \item the cognitive enhancement of language models~\citep{hollenstein2020towards, deng2023-augmented, mishra2017leveraging, long2017cognition, yang2023plm, tiwari2023predict, hollenstein2019entity, sood2020improving, takmaz2020generating, hollenstein2021reading, cheri2016leveraging, wang2024gaze, malmaud2020bridging};
    \item research on the cognitive interpretability and plausibility of language models~\citep{keller2010cognitively, beinborn2023cognitive, sood2020interpreting, bensemann2022eye, eberle2022transformer, hollenstein2021multilingual, hollenstein2022patterns, merkx2020human, hollenstein2021relative};
    \item and the investigation of text difficulty and whether it is reflected in the reading behavior, together with the provided text difficulty metrics and participants' responses to comprehension question and subjective text difficulty.
    
\end{itemize}

\noindent The \textbf{uncorrected and corrected versions of the data} can be used for 
\begin{itemize}
    \item the evaluation of rule-based algorithms that correct vertical drift in eye-tracking data~\citep{schroeder2019popeye, sanches2015eye, yamaya2017vertical, vspakov2019improving, abdulin2015person}
    \item and both the training and evaluation of machine-learning based algorithms correcting vertical drift in eye-tracking data~\citep{cohen2013software}.
\end{itemize}

\noindent The information on the different \textbf{text types} can be used for
\begin{itemize}
    \item the inference of text types from eye movement data;
    \item and research on reading behavior as a function of the type of text that is being read (poetic, argumentative, etc.), and, relatedly, investigations of the impact of different text structures and layouts on reading behavior, such as poems, dialogues, and news articles.
\end{itemize}

\noindent And last, the nature of the texts as being \textbf{machine-generated} by \textbf{different language models}, involving the information on the \textbf{decoding strategies} used by the LMs as well as the provided \textbf{transition scores, attention scores, and hidden states} can be used for
\begin{itemize}
    \item the investigation of the cognitive alignment of the output of specific LMs with humans via transition scores and attention scores;
    \item the investigation of the cognitive alignment of the output of specific decoding strategies with humans via transition scores and attention scores;
    \item the investigation of reading behavior on machine-generated texts, furthering the automated detection of machine-generated texts;
    \item the evaluation of the quality of machine-generated text with eye movements;
    \item the leveraging of eye movements to train a reward model for reinforcement learning from human feedback (RLHF);
    \item the investigation of the predictive power of surprisal estimated directly for texts that were machine-generated;
    \item and the investigation of the predictive power of surprisal when computed directly from the transition scores.
    
\end{itemize}

\section{Background}
\label{sec:background}

\subsection{Eye Movements in Reading}
\label{sec:eye-movements-in-reading}

When humans read text, their eyes perform two main actions: during \textit{fixations}, the eye remains relatively still on one point of focus and the brain obtains visual input; and during \textit{saccades}, which are ballistic relocation movements, the eye jumps from one fixation to another and is ``blind''.
While average fixations during reading last about 200-250\,ms, saccades are very rapid movements of 20-80\,ms that drive the eye about 7-9 letter spaces further in the text~\citep{rayner1998, rayner2009}. 

Reading is a complex task in which a variety of cognitive processes are involved. Different words are fixated for longer or shorter times, or they are skipped entirely~\citep{liversedge2000saccadic}. Many models have been trying either to explain specific aspects of the reading process, such as word identification or syntactic parsing, or to investigate how these subcomponents of language processing in reading (like word identification) in conjunction with other constraints (memory, visual acuity) guide readers' eyes~\citep{rayner2010models}. One of the most important models is the E-Z Reader model~\citep{reichle1998toward}, which outlines how eye movements in reading are dependent on lexical factors: features such as word frequency or predictability affect the duration of a fixation and whether or not a word is skipped. The other important model is the SWIFT model~\citep{engbert2005swift}, which posits that variables like word frequency influence fixation durations indirectly by inhibiting a random timer that determines the moment-to-moment decisions about when to initiate a saccade. 

Eye movements in reading thus reveal insight into both the cognitive processes underlying human language comprehension as well as linguistic properties of the text~\citep{radach2004theoretical}. On the one hand, they exhibit between-subject idiosyncrasies which remain consistent within-subject across different tasks~\citep{bargary2017individual} and can even be used to infer subject identity~\citep{JaegerECML2019}. On the other hand, being driven by an intricate interplay -- both voluntary and involuntary -- between cognition, attention, and oculomotor control, they can be indicative of a wide range of cognitive abilities and processes, such as cognitive load~\citep{delgado2022cognitive}, working memory~\citep{indrarathne2018role, huang2022relationship}, attention~\citep{rodrigue2015spatio}, and dyslexia~\citep{raatikainen2021detection, haller2022eye}.

\subsection{Autoregressive Language Models}
\label{sec:language-models}



Autoregressive language models (LMs), also denoted \(P_\theta\), where \(\theta\) are the model parameters, are a class of probabilistic models that assign a probability to each word in a vocabulary in order to predict the next word in a sequence conditioned on previous context.\footnote{Typically, the vocabulary is composed of a set of sub-word tokens which allow for greater flexibility in representing rare or unseen words \citep{sennrich-etal-2016-neural, kudo-richardson-2018-sentencepiece}. For simplicity, in this section we refer to these sub-word tokens as words.}
Formally, given a sequence of words as context, $\textbf{w}_{<t} \coloneqq (w_1, \ldots, w_{t-1})$, the model computes the conditional probability of all possible words $w_t$ in a predefined vocabulary $V$ at each generation step $t$:
\begin{equation}
    P_\theta(w_t | \textbf{w}_{<t}).
\end{equation}
These probabilities denote the likelihood of moving from one word to the next and are thus often referred to as transition scores.


Current state-of-the-art LMs~\citep[e.g.][]{brown2020language, touvron2023llama2} are based on decoder-only transformers~\citep{liu2018generating}. 
These comprise a stack of transformer blocks~\citep{vaswani2017attention} with a language modelling head.
The input to the model is a sequence of word embeddings representing the current context, which is passed through the layers of transformer blocks to produce a sequence of hidden states.
The final hidden state of the last context word is then passed through a linear layer to produce a score vector over the vocabulary.
These scores are unnormalised and commonly referred to as logits. 
Applying the softmax function to this set of scores converts it into a valid probability distribution over the vocabulary, where all values are positive and sum to 1.

Transformer-based autoregressive LMs can be pre-trained effectively at scale and on very large corpora using the simple self-supervised learning objective of next-word prediction.
Given a context sequence from the training corpus, the model is tasked with predicting the most likely continuation by computing.
Based on this prediction, the loss is computed as cross-entropy between the predicted distribution and the true distribution of the next word, which is backpropagated through the model to update the parameters.
Doing this repeatedly and for all possible context sequences in a large and diverse training corpus allows neural models to learn the distribution of words as well as nuanced syntactic \citep{blevins-etal-2018-deep, hewitt-manning-2019-structural, manning2020, wei-etal-2021-frequency} and semantic patterns \citep{rogers-etal-2020-primer, vulic-etal-2020-probing, zhang-etal-2021-need, merrill2024learn} in the language.

At inference time, the model is used to generate text by iteratively predicting the next word in the sequence until a special end-of-sequence (\texttt{[EOS]}) token is generated or until a maximum length is reached.

\subsubsection{Instruction-following Language Models}
\label{sec:large-language-models}

Following pre-training, the autoregressive model is typically finetuned using supervised learning techniques.
Recently, general purpose instruction tuning has emerged as a promising approach to deriving highly capable instruction-following models with impressive generalisation abilities~\citep{ouyang2022training, sanh2022multitask}.\footnote{Following pre-training and supervised finetuning, models can be further finetuned using various techniques to ``align'' the model's prediction with human preferences~\citep{ouyang2022training, rafailov2023direct, ethayarajh2024kto, azar2023general, xu2024things}.}
This step involves training the model to predict the next word in a sequence of instruction-response pairs, where the instruction is considered the context and the response is the target sequence that the model should generate.

Open-weight instruction-tuned models are available in various sizes and configurations, differing in the number of parameters and the data used for pre-training and finetuning, with numerous iterations of models being released over the past few years.
For instance, Llama~\citep{touvron2023llama2} is a family of models trained by Meta ranging from 7 billion parameters up to 70 billion parameters, pre-trained on a large-scale web corpus.
While the original Llama model was pre-trained on 1.4T tokens, Llama 2 was pre-trained on 2T tokens and subsequently finetuned on a diverse set of instruction-following tasks with supervised finetuning and reinforcement learning from human feedback (RLHF).
WizardLM~\citep{xu2023wizardlm}\footnote{\url{https://huggingface.co/WizardLM/WizardLM-13B-V1.2}} is an instruction-tuned variant of the 13-billion-parameter Llama 2 base model tuned on a diverse set of synthetic and real-world instruction-following tasks~\citep{xu2023wizardlm}.
Mistral~\citep{jiang2023mistral}\footnote{\url{https://huggingface.co/mistralai/Mistral-7B-Instruct-v0.1}} is a 7-billion-parameter model that uses a modified attention mechanism to facilitate improved efficiency and better handling of longer sequences.
There is little information available on the pre-training data used for Mistral, but the model was finetuned using publicly available instruction tuning datasets.
Phi-2~\citep{javaheripi2023phi}\footnote{\url{https://www.microsoft.com/en-us/research/blog/phi-2-the-surprising-power-of-small-language-models/}} is a 2.7-billion-parameter model pre-trained on 1.4T tokens sourced from NLP and coding datasets available on the web as well as high-quality synthetic data. 
Despite not having undergone instruction tuning, this model is still adept at following instructions and responding to prompts, presumably due to the nature of its pre-training data.
Given adequate instruction, these models are capable of producing high-quality short snippets of text that in most cases are indistinguishable from human-written texts.

\subsubsection{Decoding Strategies}
\label{sec:decoding-strategies}

To generate text from an autoregressive LM, a decoding strategy is used to select which word is generated next given the probability distribution over all words in the vocabulary computed by the LM at each timestep.
Numerous decoding strategies have been proposed, each with their own trade-offs in terms of speed and quality, as well as inherent biases that they impose on the generation process \citep{holtzman2019curious, meister-etal-2020-beam, eikema-aziz-2020-map}.
The most popular strategies can be placed into two categories: likelihood-maximization methods, which encompasses greedy search and beam search; and stochastic methods, which involves ancestral sampling, top-$k$ sampling, and top-$p$ sampling.


\paragraph{Likelihood-Maximization Strategies}

Likelihood-maximization strategies aim to approximate maximum \textit{a posteriori} (MAP) search over all possible output sequences conditioned on the context under the model. 
This approximation is necessary since exact MAP decoding is intractable given the exponentially large search space for any given text sequence.

\textbf{Greedy decoding} is a relatively simple method that selects the word \(w_{t}\) out of the vocabulary \(V\) with the highest probability at each generation step \(t\):

\begin{equation}
w_{t} = \argmax_{w \in V} P_{\theta} (w\mid\textbf{w}_{<t}).
\end{equation}

While fast, greedy decoding has been found to result in dull and repetitive texts, often exhibiting degenerate word repetitions and other disfluencies~\citep{holtzman2019curious}.

\textbf{Beam search} is a more sophisticated strategy that aims at alleviating the issue of making locally optimal decisions in constructing a sequence by maintaining multiple possible hypotheses.
At each generation step, each hypothesis is expanded by considering the $l$ most likely words, resulting in $l^2$ possible continuations (\emph{i.e.}, \textit{beams}).
Hypotheses are then pruned to keep only the $l$ most likely sequences and the process is repeated until all hypotheses have reached the end-of-sequence token or a maximum length, at which point the sequence with the highest probability is returned as the final output.\footnote{When $l = 1$, beam search is equivalent to greedy search.}
More formally, consider the $l$ solutions $S_{t-1}$ for beam search at time step~$t-1$: $S_{[t-1]}=\left\{\mathbf{w}_{<t,1}, \ldots, \mathbf{w}_{<t, l}\right\}$, where $\mathbf{w}_{<t,k}$ are the top $l$ highest scoring continuations. At time step $t$, beam search considers all possible continuations $w\in V$ and retains the $l$ highest, i.e. \begin{align}
    S_t &= \argmax_{w_1, \ldots, w_l\in V}\sum_{i\in[l]}P_\theta(\mathbf{w}_{<[t+1],i})\\
    &= \argmax_{w_1, \ldots, w_l\in V}\sum_{i\in[l]}\sum_{w_i\in V}P_\theta(w_i\mid\mathbf{w}_{<t,i}).
\end{align}
While beam search has long been the de facto choice of decoding strategies in tasks such as machine translation and speech recognition~\citep{och-etal-1999-improved, koehn-etal-2003-statistical}, it often tends to produce less diverse outputs~\citep{vijayakumar2018diverse}, making it less suitable for more creative, open-ended style generation tasks. 
Additionally, it is slower and more memory intensive than other decoding strategies due to having to keep track of multiple hypotheses.

\paragraph{Stochastic Strategies}

In order to improve the diversity of generated text, a number of strategies rely on stochastic sampling from the model's output distribution.

\textbf{Ancestral sampling} randomly draws $w_t$ from the model's full probability distribution over the vocabulary \(V\) at each generation step:

\begin{equation}
{w}_{t} \sim P_{\theta}(w\mid\textbf{w}_{<t}).
\end{equation}

While this sampling strategy allows for constructing a text sequence that is faithful to the model, it is susceptible to errors and biases present in the estimated distribution, particularly in the long tail of low-probability words~\citep{freitag-etal-2023-epsilon}.
To mitigate this, other sampling strategies truncate the output distribution to restrict the selection to high-probability words.

\textbf{Top-$k$ sampling}~\citep{fan-etal-2018-hierarchical} considers only the $k$ most likely candidates at each time step: the vocabulary \(V\) is subset to \(V_{k,t}\) such that \(V_{k,t}\) consists of the \(k\) most likely words at the current generation step \(t\):

\begin{equation}
    V_{k,t} = \underset{V' \subset V: |V'| = k }{\argmax} \hspace{.5em}\underset{w \in V'}{\sum} P_\theta\left(w\mid\textbf{w}_{<t}\right).
\end{equation}

Here, $V' \subset V: |V'| = k $ denotes all possible subsets of the vocabulary $V$ with cardinality $k$. 
$w_{t}$ is then drawn from the renormalized top-$k$ probability distribution for tokens in \(V_{k,t}\) and 0 otherwise:

\begin{equation}
    P(w_t\mid\textbf{w}_{<t}) = \begin{cases}
    \frac{P_{\theta}(w\mid \textbf{w}_{<t})}{\underset{w \in V_{k,t}}{\sum} P_\theta(w\mid\textbf{w}_{<t})} & \text{if } w \in V_{k,t}, \\
    0 & \text{otherwise}.
    \end{cases}
\end{equation}

In this strategy, $k$ is set to a fixed value between $\left[1, \left|V\right|\right]$, with popular values typically ranging from 10 to 50, depending on the task.\footnote{Note that $k = 1$ is equivalent to greedy search while $k = \left|V\right|$ is equivalent to ancestral sampling.
}

\textbf{Top-$p$} (or \textbf{nucleus}) \textbf{sampling}~\citep{holtzman2019curious} constructs the minimal subset of possible candidates by keeping only those words whose cumulative probability mass is above a certain threshold $p$.
This subset denotes the \textit{nucleus} and often contains anywhere between 1 and 1000 words~\citep{holtzman2019curious}.
Thus, in contrast to top-$k$, the size of the considered subset is dynamic given the model's output distribution at each timestep. The vocabulary \(V\) is subset to the nucleus \(V_{p,t}\), formalized as 

\begin{equation}
    V_{p,t}=min\left|\left\{w\in V: \sum_{w\in V}P_\theta(w\mid\mathbf{w}_{<t})>p\right\}\right|.
\end{equation}

Again, $w_{t}$ is then drawn with probability proportional to the original probability
distribution for tokens in the nucleus $V_{p,t}$ and 0 otherwise:

\begin{equation}
    P(w_t\mid\textbf{w}_{<t}) = \begin{cases}
    \frac{P_{\theta}(w\mid\textbf{w}_{<t})}{\underset{w \in V_{p,t}}{\sum} P_\theta(w\mid\textbf{w}_{<t})} & \text{if } w \in V_{p,t}, \\
    0 & \text{otherwise}.
    \end{cases}    
\end{equation}



Temperature scaling can also be used to modify the sharpness of the model's next-token probability distribution when computing it over the set of output logits $z_i$


\begin{equation}
        q_i = \frac{\text{exp}(z_i / \tau)}{\sum_{j=1} \text{exp}(z_j / \tau)}.
\end{equation}

Higher temperatures ($\tau > 1$) flatten the distribution and increase stochasticity for all sampling strategies, while lower temperatures ($\tau < 1$) sharpen the distribution, effectively decreasing stochasticity.

\section{Related Work}
\label{sec:related-work}

The following section outlines an overview on existing corpora of eye movements in reading. Most of the existing eye movements in reading datasets consist of minimal pair stimuli; we will only review corpora comprising naturalistic or at least semi-naturalistic (semi-constructed) stimuli that are multi-purpose and do not follow a minimal pair design. For a more detailed review on available corpora, please refer to \citet{potec}, who provide an in-depth survey.

Existing corpora of eye movements in reading can be divided according to the text presented, \emph{i.e.}, whether they are single-sentence or entire paragraphs, and according to whether they are mono- or multilingual.

\subsection{Single-Sentence Corpora of Eye Movements in Reading}
\label{sec:rel-work-single-sentence}

Eye-movement corpora containing data of subjects reading single sentences can be categorized into those corpora that comprise \textit{partially constructed stimuli} and those comprising \textit{naturalistic stimuli}. Partially constructed stimuli typically concentrate on linguistic constructions that might not occur frequently enough in naturalistic language examples and are thus authored for the purpose of studying different phenomena. For instance, the Potsdam Sentence Corpus~\citep{kliegl2004psc} consists of 144 German sentences that entail a variety of linguistic structures, read by 33 younger 32 older adults. Other single-sentence corpora are TURead~\citep{Acartrk2023} with 196 participants reading Turkish stimuli; The Potsdam-Allahabad Hindi Eyetracking Corpus~\citep{Husain-Vasishth-Srinivasan-2014-hindi}, containing 30 participants reading 153 sentences in Hindi or Urdu; a Chinese dataset with the purpose of studying word length effects, comprising 30 participants reading 90 Chinese sentences~\citep{Zang2018zh-word-len}; and the Russian Sentence Corpus~\citep{Laurinavichyute2018RSC}, with 96 participants reading 144 sentences.

Naturalistic single-sentence corpora, on the other hand, involve the presentation of sentences that occur in natural language. The vast majority of these corpora is based on English texts. 
The Corpus of Eye Movements in L1 and L2 English Reading (CELER)~\citep{celer2022} is the largest corpus in terms of number of participants, with a total of 365 subjects
(roughly 80\% of which are non-native speakers) 
reading 156 English sentences from the Wall Street Journal. 
As another example, The Zurich Cognitive Language Processing Corpora, ZuCo~\citep{Hollenstein2018zuco1} and ZuCo 2~\citep{hollenstein-etal-2020-zuco}, comprise 12 and 18 native English speakers respectively, who were concomitantly tracked with EEG. 
Other corpora include the UCL corpus~\citep{frank2013reading} with 43 participants reading sentences from English novels; RaCCooNS~\citep{Frank2023} with 37 subjects reading 200 narrative sentences; the CFILT sarcasm dataset~\citep{Mishra-Kanojia-Bhattacharyya-2016}, entailing 7 non-native English speakers reading sarcastic and non-sarcastic sentences; and the CFILT sentiment complexity dataset~\citep{joshi-etal-2014-measuring}, where 5 participants rated the sentiment of over 1000 English sentences. Chinese single-sentence corpora involve the Hong Kong Corpus of Chinese Sentence and Passage Reading~\citep{Wu2023HKC}, including 96 participants that read 300 sentences and seven multi-line passages; the Beijing Sentence Corpus~\citep{Pan2021BSC}, with 60 participants reading 150 sentences from People's Daily; and \citet{Zhang2022chinese} tracked 1718 subjects reading over 7500 Chinese sentences.

\subsection{Naturalistic Text Passage Corpora of Eye Movements in Reading}
\label{sec:rel-work-text-passage}

Naturalistic reading entails the reading of texts that occurs or could occur naturally in the language. Naturalistic stimuli do not contain experimentally manipulated syntactic structures or lexical items that target a specific research hypothesis, as opposed to constructed or partially constructed stimuli. Naturalistic reading does not include task solving, time constraints or other constraints such as restricting the preview of the subsequent word or phrase.

\paragraph{Monolingual Corpora}

Eye movement data in reading naturalistic stimuli 
already exists for some languages, predominantly for English. For instance, the Provo corpus~\citep{luke2018provo} contains the data of 84 participants reading 55 English texts from a variety of sources; GazeBaze~\citep{griffith2021gazebase} provides the data of 322 subjects reading up to 18 passages of a poem over the course of three years; the Individual Differences Corpus (InDiCo)~\citep{haller_measurement_2023} entails German naturalistic reading data from four experimental sessions from each participant, including a battery of psychometric test scores; PoTeC~\citep{potec} consists of data from German native speakers, who can be categorized into levels of expertise, reading German textbook passages from two different domains (physics and biology); and MECO-L2~\citep{kuperman2023text} contains 543 non-native English speakers of 12 different L1 backgrounds reading 12 English texts. 
Other monolingual corpora include the PopSci Corpus~\citep{wolfer2013popsci} (17 participants reading 16 German popular science texts); OneStopQA~\citep{malmaud2020bridging} (296 participants reading 10 articles, with comprehension questions presented either before or after reading the text); RastrOS~\citep{Leal2022rastros} (37 subjects reading 50 paragraphs in Portuguese; the Mental Simulation Corups~\citep{mak2019mentalsimulation} (102 participants reading three literary short stories in Dutch and subsequently answering questions); and a corpus targeted at studying parafoveal processing~\citep{parker2017predictability}, placing target words either inside or outside of parafoveal view.

Some naturalistic eye-tracking-while-reading corpora are designed with the purpose of inferring reader- or text-specific properties from the data or of being employed for NLP use cases. They usually entail annotation layers that either categorize the subjects or the texts, or both. For instance, SB-SAT~\citep{ahn2020towards} consists of data from 95 participants reading 4 passages from the Scholastic Assessment Test (SAT) while also rating their subjective difficulty. MAQ-RC~\citep{sood2020improving} had 23 participants reading 32 movie plots and answering comprehension questions. Other corpora include the CFILT co-reference dataset~\citep{cheri2016leveraging}, the CFILT scanpaths dataset~\citep{mishra2018scanpath}, the CFILT text quality dataset~\citep{mathias-etal-2018-eyes}, and the CFILT essay grading dataset~\citep{mathias2020happy}.

\paragraph{Multilingual Corpora}

There also exists a range of multilingual eye-tracking-while-reading corpora on the passage level. The Multilingual Eye-Movements Corpus (MECO-L1)~\citep{siegelman2022expanding} is among the biggest ones, being a 13-language reading corpus with 45-55 subjects per language who read 12 encyclopaedic texts. The Ghent Eye-Tracking Corpus (GECO)~\citep{cop2017presenting} had 14 English monolinguals read an entire novel in English and 19 bilinguals read half of the novel in English and the other half in Dutch. Its extension, GECO-CN~\citep{sui2023geco} employed the same stimulus texts, but the subjects were 30 Chinese native speakers reading half of the novel in Chinese and the other half in English. The Copenhagen Corpus of Eye Tracking Recordings (CopCo)~\citep{hollenstein-etal-2020-zuco} comprises 22 native Danish speakers, 19 dyslexic speakers, and 10 Danish L2 speakers reading 20 speech manuscript in Danish; and CopCo L2~\citep{reich-etal-2024-reading-equal} then provided an extension with L2 recordings. Other multilingual eye-tracking-while-reading corpora include WebQAmGaze~\citep{ribeiro2023webqamgaze} (German, English, Spanish and Turkish recordings with a webcam); Dundee~\citep{kennedy2003dundee} (10 English and 10 French native speakers reading newspaper extracts); a corpus of 50 participants reading 100 Chinese articles with the task of summarization upon reading~\citep{yi2020dataset}; SummarEyes~\citep{summareyes} (80 participants read and summarized 4 different texts); and a corpus of 29 subjects judging the relevance of a given document for a given query of 60 document-query pairs~\citep{li2018understanding}.

\subsection{Corpora of Eye Movements on Machine-Generated Texts}
\label{sec:rel-work-eye-movs-machine-generated-texts}

To the best of our knowledge, there exists only a single corpus that comprises eye-tracking-while-reading data on machine-generated texts: the Ghent Eye-Tracking Corpus of Machine Translation (GECO-MT)~\citep{colman-etal-2022-geco}. The purpose of this corpus is to compare human and machine translations, and it contains eye movement data on both a human and a machine translation of Agatha Christie's novel \textit{The Mysterious Affair at Styles}. 20 Dutch-speaking university students participated in the experiment. The machine translations were generated with Google Translate, deployed in May 2019. They do not provide any model-related information or model internals, which is an inherent limitation of using closed-source services.

\section{Stimuli}
\label{sec:stimuli}

This section outlines the generation of the stimuli including the steps taken to post-process and annotate them with linguistic annotation layers. 
Additionally, we also describe the generation of the comprehension questions related to the stimuli and provide descriptive statistics of the experimental items. 
All stimuli can be found in the file \texttt{data/stimuli.csv}, which not only contains the stimuli, the prompts, the comprehension questions and possible answers as presented in the eye-tracking experiment but also all stages of post-processing, including unique word and token identifiers that allow to map between sub-word tokens and word tokens and between sub-word tokens and the sub-word token-level transition scores, attention scores, and hidden states. A detailed description of all columns in the stimuli file is provided in file \texttt{data/stimuli\_columns\_descriptions.csv}.

\subsection{LLMs for Stimuli Generation}
\label{sec:stimulus-generation-prompting-llms}

To generate the reading stimuli, we use three large language models (LLMs) of different size and from different language model families: Phi-2~\citep{javaheripi2023phi}; Mistral 7B Instruct~\citep{jiang2023mistral}; and WizardLM 13B~\citep{xu2023wizardlm} (for more details, refer back to Section~\ref{sec:large-language-models}). 
In selecting LLMs for this task, we refrain from using models that have undergone extensive alignment stages such as reinforcement learning from human feedback (RLHF) \citep{ouyang2022training}. 
RLHF is a policy-based approach which involves generating candidate responses to an input prompt ({i.e.}, referred to as the rollout), scoring the generated responses with a reward model and then updating the policy (in this case, the model) accordingly.
Related work does not explicitly state which decoding strategies are used during the rollout phase \citep{touvron2023llama2, ouyang2022training, stiennon2020learning}.
However, we hypothesise that the selection of a certain decoding strategy used during policy-based training may introduce potential biases towards certain types of decoding strategies.
To mitigate this, we avoid using models trained with RLHF.
Instead, we restrict our model selection to pre-trained and instruction-tuned models, which are trained with a standard maximum likelihood estimation (MLE) objective. 
Specific reasons for choosing the three models listed above as opposed to other models are outlined in Appendix~\ref{appendix:model-choice}.

For each model, we generate five output texts for a given prompt by employing the five different decoding strategies described in \S \ref{sec:decoding-strategies}: greedy search, beam search\footnote{At the time of model prompting, beam search was not implemented for Phi-2.}, ancestral sampling, top-$k$ sampling~\citep{fan-etal-2018-hierarchical}, and top-$p$ sampling~\citep{holtzman2019curious}. 

\subsubsection{Prompts}
\label{sec:prompts}

Our prompts are designed to cover different tasks and distinct text types. 
This ensures that the output is as varied as possible to allow for a high generalizability of the analyses that can be done with the corpus. The two main generation tasks are \textit{unconstrained} and \textit{constrained} text generation: during the \textit{unconstrained} generation, the model is not restricted with respect to its output, meaning it can be neither ``right'' nor ``wrong''. 

The \textit{unconstrained} prompts are further divided into \textit{non-fiction}, prompting the model to either write an argument on a certain topic or provide a description of a topic or scene; \textit{fiction}, prompting the model to either write a short story or a dialogue between two characters; and \textit{poetry}, prompting the model to compose a poem. 

In the \textit{constrained} setting, the model was restricted pertaining to its output, \emph{i.e.}, it was clear to some extent what the model was expected to produce and there is a way for the model to hallucinate (when the model produces output that sounds plausible but is either fabricated or inaccurate). \textit{Constrained} prompts can be further divided into \textit{summarization}, prompting the model to provide a summarization of an input text; \textit{synopsis}, where the model is asked to write a news article based on a synopsis it is provided with; and \textit{key-word text generation}, comprising the model being given five key words and being prompted to compose a text based on them. An overview of the different prompt types can be seen in Table~\ref{tab:prompt-categories}, and the full list of prompts can be found in the file \texttt{data/stimuli.csv}. 

\begin{table}[h]
\caption{The prompt types used and the number of prompts.}
\label{tab:prompt-categories}
\small
\centering
\begin{tabular}{llllp{4cm}}
\hline
Type                           & Task                         & Sub-Type    & Number & Description                                                                          \\ \hline
\multirow{5}{*}{Unconstrained} & \multirow{2}{*}{Non-Fiction} & Argument    & 6      & The generated text should present an argument on a certain topic.                    \\ 
                               &                              & Description & 5      & The generated text describes a topic or scene.                                       \\   \hhline{~----} 
                               & \multirow{2}{*}{Fiction}     & Story       & 3      & The model is prompted to write a story.                                              \\
                               &                              & Dialogue    & 3      & The model is prompted to write a dialogue between two characters.                                          \\ \hhline{~----} 
                               & Poetry                       &             & 4      & The model is prompted to write a poem.                                               \\ \hline 
\multirow{3}{*}{Constrained}   & Summarization                &             & 8      & The model is asked to generate a summary of a provided text.                                  \\  \hhline{~----} 
                               & Synopsis                     &             & 7      & The model is asked to write a news article based on a synopsis. \\   \hhline{~----} 
                               & Key-word text                  &             & 6      & The model is given key words and has to compose a text based on them.                      \\ \hline
\end{tabular}
\end{table}

Instruction-tuning typically involves formatting training examples with a pre-defined prompt template. 
These prompt templates often differ across models, as shown in Figure~\ref{fig:prompt-templates}.
In order to construct the full prompts used to generate the stimuli from each model, we fill the model-specific prompt templates accordingly.

\paragraph{Item Selection}

Our corpus consists of texts generated from 42 different prompts, which were selected based on criteria outlined further below. To conduct the selection, we crafted a total of 106 prompts for which the models generated outputs with all investigated decoding strategies. From the model outputs, we then selected 42 items that resulted from 42 prompts, each item consisting of 14 conditions, one for each combination of model and decoding strategy. The selection criteria ensured that all the 14 conditions corresponding to a single item were suitable to be presented to the participants in an eye-tracking experiment who were unaware of the fact that the stimuli were machine-generated. Items were excluded if for at least one condition the model i) responded only with one word or one short sentence ii) the model repeated the question or the prompt iii) the model was self-referential (\emph{e.g.,} \textit{As an AI model, ...}), or iv) there was an \textit{I}-narrator. We further selected the items in a way that the different text types are balanced.

\subsubsection{Generation Parameters and Experiment}
\label{sec:generation parameters}

When generating texts with language models, there is a range of hyper-parameters that have to be determined \textit{a priori} which control the generation process and shape the model output. In order to be able to present one text on a single screen during the eye-tracking experiment (as opposed to splitting it across several screens) while maintaining a font size that is not only legible but also big enough to account for a facilitated attribution of fixations to areas of interest~\citep{holmqvist2011eye}, we set the maximal number of generated tokens to 150\footnote{Number of tokens refers to sub-word tokens, not words.}. For sampling-based decoding strategies, we used a temperature of $0.8$. We set $k=50$ for top-$k$ sampling, and $p=0.9$ for top-$p$ sampling.
These settings aim to allow for a sufficient degree of creativity while maintaining coherency and fluency.
For beam search, we set the number of beams ($l$) to 4 and did not apply early stopping. 
We refrained from using larger beam sizes due to the known deficiencies of beam search \citep{koehn-knowles-2017-six}, which relate an increase in beam size to a decrease in output quality, as well as the additional computational overhead involved with keeping track of a larger number of hypotheses.
For a detailed account on the decoding parameters, please refer back to Section~\ref{sec:decoding-strategies}.

We loaded the models across five NVIDIA GeForce RTX 3090 GPUs for generation. All models are implemented in PyTorch~\citep{pytorch2019paszke} and loaded from Huggingface~\citep{wolf2019huggingface}.

\subsubsection{Post-Processing of Generated Stimuli Texts}
\label{sec:post-processing-generated-texts}

\paragraph{Truncation}

In order to present participants only with complete texts so as not to bias their eye movements, we removed incomplete sentences from the generated model outputs, resulting from the models attempting to exhaust the full 150 tokens set as maximum number of tokens to generate. We further removed \textsc{EOS} tokens and trailing white-space and newline characters. Regarding poems, we additionally restricted the number of newline characters to 9 such that the poems fit onto the presentation screen, with poems usually consisting of 2 stanzas, 4 verses each and a newline in-between. The final texts as they were presented in the eye-tracking experiments can be found in the column \texttt{gen\_seq\_trunc} in the \texttt{stimuli.csv} file.

\paragraph{Cleaning the Data}
Having prospective analyses with eye movement data in mind, we post-processed the generated sequences and also release the different stages of post-processing. With the fixation data and the reading measures being on word-level, \emph{i.e.}, areas of interest to which fixations and reading measures are mapped to are obtained by splitting a text on its white-space characters, data cleaning ensures that this mapping to word-level can be executed correctly.

Data cleaning to ensure suitability for eye movement data analyses involves removing trailing and superfluous white-space characters, punctuation marks, accounting for otherwise problematic output and ultimately mapping the sub-word level model output to word-level. The data cleaning process consisted of the following pipeline: 
\begin{enumerate}
    \item Map the generated sub-word tokens to word ids. This ensures that sub-word tokens belonging to the same word are grouped together and that transition and attention scores and hidden states can be mapped to the correct word.\footnote{Within the eye movements in reading research community, it has been a standard approach to define word-based interest areas, as even high-precision eye-trackers are not always accurate to the character-level.} This allows for a proper one-to-one mapping between the words and the areas of interest as defined in the eye-tracking experiment. 
    \item Remove newlines and white-space characters, as they do not obtain fixations. This is done by attributing a unique token id to every generated token, which also includes white-space characters, and then subsetting these token identifiers to those which will potentially have fixations on them. These token identifiers can then in turn be used to subset the transition scores, attention scores, and hidden states and discard those that do not refer to an area of interest. 
    \item We optionally included a further stage of post-processing that allows for removing sub-word tokens and scores referring to punctuation marks to allow for the possibility of not including the scores the language models produced for punctuation when computing word-level metrics such as surprisal or entropy.
\end{enumerate}

\noindent The post-processing steps merely result in the correct subset of unique token identifiers that enable a correct mapping between areas of interests that contain fixations and reading measures and the generated tokens in case of analyses with the scores; they allow for accessing not only the original generated sub-word tokens and scores before post-processing, but also the truncated versions, the versions without white-space characters and newlines, and the versions without punctuation marks. The experimental stimuli presented in the experiment are acquired from the first step of post-processing, which is truncation; all white-space characters, newlines and punctuation marks are thus presented to the participants as the language models generated them.

\subsection{Lexical Annotation}
\label{sec:lexical-annotation}

\paragraph{Annotation on Text Level}
 On the text level, we report the length of a text in number of words, the average word length in number of characters, the average Zipf frequency and the average word frequency obtained from the \texttt{wordfreq} library\footnote{\url{https://pypi.org/project/wordfreq/}}. The frequency of a word is reported for words that appear at least once in 100 million words as a min-max normalized decimal between 0 and 1. The Zipf frequency of a word is the word's base-10 logarithm of the number of time is appears in a billion words. On text level, the word frequency and the Zipf frequency values are averaged across all words in a text. 

 We further report eight different text readability metrics computed with the \texttt{readability} package\footnote{\url{https://pypi.org/project/py-readability-metrics/}}. They include the Flesch Reading Ease and the Flesch Kincaid Grade Level~\citep{kincaid1975derivation}; the Gunning Fog Index~\citep{gunning2004plain}; the Coleman-Liau Index~\citep{coleman1975computer}; the Dale-Chall Readability Formula~\citep{dale1948formula}; the Automated Readability Index~\citep{senter1967automated}; the Linsear-Write Readability Metric~\citep{o1966gobbledygook}; and the Spache readability formula\citep{spache1953new}. Note that these metrics can only be computed above a certain threshold text length.

 \paragraph{Annotation on Word Level}
 On word level, we report the word length in characters with and without punctuation; the part-of-speech, the dependency tag, the number of left and right dependents, and the distance to the head, all extracted with \texttt{spaCy}~\footnote{\url{https://spacy.io/}}; whether or not a word is last in line as presented in the eye-tracking experiment; the word frequency and the Zipf frequency, again computed with the \texttt{wordfreq} library; and the surprisal~\citep{hale2001probabilistic, levy2008expectation}, which is the negative log-probability of a word given its preceding context. Since the predictive power of surprisal on human reading times has been shown to differ depending on the language model from which it has been extracted ~\citep{goodkind2018predictive, oh-schuler-2023-transformer, oh2023does, wilcox-etal-2023-language, wilcox2020predictive}, we estimate surprisal from a range of different language models: GPT-2 \textit{base} and \textit{large}~\citep{radford2019language}; OPT-350m and OPT-1.3b~\citep{zhang2022opt}; Llama-2-7b and Llama-2-13b~\citep{touvron2023llama2}; Phi-2~\citep{javaheripi2023phi}; Mistral 7b~\citep{jiang2023mistral}; and Pythia-6.9b and Pythia-12b~\citep{biderman2023pythia}. Moreover, we estimate surprisal and entropy for the stimulus text in two different ways: once the input to the above-mentioned language models is the generated stimulus text alone, and once the input is both the combination of prompt and generated stimulus text. The latter is provided for the option of having a consistent comparison with the transition scores, which are conditioned on the prompt as well.

\subsection{Comprehension Question Generation}
\label{sec:comprehension-question-generation}
For each of the 42 items in each of the 14 experimental conditions, we created a multiple-choice comprehension question with four possible answers. The purpose of these questions is to encourage participants to maintain their attention level and properly read the text, as opposed to merely skimming them and mindlessly clicking through the experiment. The comprehension questions were generated by prompting ChatGPT~\citep{brown2020language, achiam2023gpt}. The prompt template used is provided in Appendix~\ref{appendix:comprehension-question-generation}.

To ensure the quality and correctness of the questions' content, we manually validated all generated questions and answers and made minor edits to instances where any of the following issues were observed: 
\begin{enumerate*}[label=\roman*)]
\item more than one possible answer was true; 
\item the length of the true answer differed from the length of the three false answers; 
\item the question included an entire phrase of the text verbatim; or
\item the question did not refer to the text directly but to world knowledge about the topic of the text. 
\end{enumerate*}
We further re-attributed the positions of correct answers such that their ultimate distribution (\emph{i.e.}, whether the correct answer was answer a, b, c, or d) is uniform.

\subsection{Descriptive Statistics of the Generated Stimuli}
\label{sec:descriptive-statistics-stimuli}

\begin{table}[]
\caption{
Descriptive statistics across all experimental stimuli in \emph{all conditions}.}
\label{tab:stats-text-overall}
\begin{tabular}{llllll}
\hline
     & Text Length & Word Length & Zipf Freq. & Flesch  & Surp. GPT-2 \\ \hline
mean & 86.515      & 5.16        & 5.644      & 43.035  & 4.018       \\ 
std  & 20.675      & 0.519       & 0.229      & 22.631  & 3.77        \\ 
min  & 25.0        & 3.621       & 4.827      & 4.783   & 0.0         \\ 
max  & 130.0       & 6.66        & 6.406      & 108.073 & 30.897      \\ \hline
\end{tabular}
\end{table}

We provide the descriptive statistics of a selection of the lexical characteristics of the stimuli. Table~\ref{tab:stats-text-overall} depicts the overall average text length, the average word length, the average Zipf frequency, the average Flesch reading ease score, and the average surprisal value extracted from GPT-2 \textit{base}. 

\begin{table}[]
\caption{
Descriptive statistics for \emph{each model}.
}
\label{tab:stats-text-models}
\begin{tabular}{lllllll}
\hline
                          &      & Text Length & Word Length & Zipf Freq. & Flesch  & Surp. GPT-2 \\ \hline
\multirow{4}{*}{Phi-2}     & mean & 84.327      & 5.131       & 5.686      & 40.228  & 4.0         \\
                          & std  & 26.246      & 0.571       & 0.217      & 25.755  & 3.728       \\
                          & min  & 25.0        & 3.795       & 4.889      & 11.623  & 0.0         \\
                          & max  & 130.0       & 6.411       & 6.189      & 105.304 & 30.897      \\ \hline
\multirow{4}{*}{Mistral}  & mean & 89.995      & 5.155       & 5.643      & 44.775  & 3.935       \\
                          & std  & 19.287      & 0.462       & 0.234      & 20.18   & 3.714       \\
                          & min  & 37.0        & 3.621       & 4.943      & 5.949   & 0.0         \\
                          & max  & 124.0       & 6.66        & 6.406      & 100.731 & 30.897      \\ \hline
\multirow{4}{*}{WizardLM} & mean & 84.786      & 5.189       & 5.611      & 43.54   & 4.121       \\
                          & std  & 16.097      & 0.53        & 0.23       & 22.172  & 3.859       \\
                          & min  & 28.0        & 3.903       & 4.827      & 4.783   & 0.0         \\
                          & max  & 118.0       & 6.405       & 6.117      & 108.073 & 30.897      \\ \hline
\end{tabular}
\end{table}

The same statistics can be found in Table~\ref{tab:stats-text-models}, where they are averaged not across all stimuli but across those stimuli generated by one LLM. Mistral produces, on average, the longest texts, and while WizardLM and Phi-2 are equal with respect to text length, Phi-2 exhibits the greatest variability of all three models. According to the Flesch score, Phi-2 also appears to generate, on average, slightly easier texts than the other two models, although its standard deviation is again highest. 

\begin{table}[]
\caption{
Descriptive statistics for each \emph{generation task}.
}
\label{tab:stats-text-text-types}
\begin{tabular}{lllllll}
\hline
                               &      & Text Length & Word Length & Zipf Freq. & Flesch  & Surp. GPT-2 \\ \hline
\multicolumn{7}{c}{\textit{Unconstrained}} \\ \hline    
\multirow{4}{*}{Non-fiction}   & mean & 96.247      & 5.276       & 5.706      & 41.635  & 3.703       \\
                               & std  & 16.071      & 0.412       & 0.182      & 15.482  & 3.378       \\
                               & min  & 44.0        & 3.901       & 5.139      & 12.497  & 0.0         \\
                               & max  & 127.0       & 6.217       & 6.189      & 91.995  & 27.205      \\ \hline
\multirow{4}{*}{Fiction}       & mean & 93.536      & 4.608       & 5.806      & 72.119  & 3.598       \\
                               & std  & 16.315      & 0.398       & 0.19       & 17.238  & 3.473       \\
                               & min  & 42.0        & 3.853       & 5.178      & 46.583  & 0.0         \\
                               & max  & 130.0       & 5.59        & 6.086      & 108.073 & 30.897      \\ \hline
\multirow{4}{*}{Poetry}        & mean & 57.518      & 4.493       & 5.628      & 50.667  & 5.743       \\
                               & std  & 11.623      & 0.365       & 0.311      & 33.947  & 4.412       \\
                               & min  & 37.0        & 3.621       & 4.943      & 51.959  & 0.0         \\
                               & max  & 84.0        & 5.25        & 6.406      & 91.445  & 30.897      \\ \hline
\multicolumn{7}{c}{\textit{Constrained}} \\ \hline   
\multirow{4}{*}{Summarization} & mean & 78.071      & 5.28        & 5.538      & 40.199  & 4.549       \\
                               & std  & 19.573      & 0.36        & 0.22       & 18.882  & 4.103       \\
                               & min  & 28.0        & 4.514       & 4.827      & 18.176  & 0.0         \\
                               & max  & 113.0       & 6.303       & 5.976      & 77.069  & 24.651      \\ \hline
\multirow{4}{*}{Synopsis}      & mean & 88.878      & 5.48        & 5.51       & 27.874  & 4.119       \\
                               & std  & 17.407      & 0.376       & 0.184      & 11.435  & 3.99        \\
                               & min  & 40.0        & 4.5         & 5.016      & 5.949   & 0.0         \\
                               & max  & 128.0       & 6.66        & 5.876      & 59.635  & 29.789      \\ \hline
\multirow{4}{*}{Key words}   & mean & 89.488      & 5.413       & 5.677      & 32.898  & 3.604       \\
                               & std  & 20.497      & 0.409       & 0.188      & 14.306  & 3.541       \\
                               & min  & 25.0        & 4.649       & 5.259      & 4.783   & 0.0         \\
                               & max  & 124.0       & 6.411       & 6.048      & 62.19   & 29.315      \\ \hline
\end{tabular}
\end{table}

In Table~\ref{tab:stats-text-text-types}, the same lexical characteristics are displayed across text types. As to be expected, we observe a high degree of variability within the \textit{unconstrained} text types, with \textit{non-fiction} and \textit{fiction} producing the longest texts and \textit{poetry} the shortest. The \textit{constrained} text types' average text lengths are closer together in magnitude, with \textit{summarization} producing, on average, the shortest texts, probably owing to the nature of the task. There are also discernible differences pertaining to Flesch reading ease: the \textit{fiction} texts are, on average, the most difficult ones, followed by \textit{poetry}, which exhibits the highest variability in difficulty; the articles generated based on an article \textit{synopsis} are on average the easiest, according to the Flesch score. Regarding surprisal, the \textit{poetry} texts contain, on average, the words with the highest surprisal, as opposed to \textit{fiction}, with the lowest average surprisal score, followed by the texts composed based on \textit{key words}. 
Overall, these statistics align with our expectations for the distinct text types.

\begin{table}[]
\caption{
Descriptive statistics for each \emph{decoding strategy}.
}
\label{tab:stats-text-dec-strategies}
\begin{tabular}{lllllll}
\hline
                               &      & Text Length & Word Length & Zipf Freq. & Flesch  & Surp. GPT-2 \\ \hline
\multirow{4}{*}{Greedy search} & mean & 85.77       & 5.11        & 5.675      & 44.542  & 3.914       \\
                               & std  & 20.322      & 0.526       & 0.229      & 22.849  & 3.698       \\
                               & min  & 40.0        & 3.621       & 4.951      & 7.097   & 0.0         \\
                               & max  & 126.0       & 6.032       & 6.406      & 103.438 & 30.897      \\ \hline
\multirow{4}{*}{Beam search}   & mean & 86.714      & 5.152       & 5.629      & 44.522  & 3.945       \\
                               & std  & 17.479      & 0.511       & 0.23       & 22.199  & 3.779       \\
                               & min  & 44.0        & 3.903       & 4.827      & 8.631   & 0.0         \\
                               & max  & 118.0       & 6.012       & 6.045      & 103.673 & 30.897      \\ \hline
\multirow{4}{*}{Sampling}      & mean & 85.817      & 5.208       & 5.629      & 41.517  & 4.116       \\
                               & std  & 20.824      & 0.524       & 0.233      & 22.381  & 3.836       \\
                               & min  & 34.0        & 3.853       & 4.889      & 5.949   & 0.0         \\
                               & max  & 127.0       & 6.66        & 6.117      & 108.073 & 30.897      \\ \hline
\multirow{4}{*}{Top-$k$}       & mean & 87.635      & 5.165       & 5.641      & 43.872  & 4.088       \\
                               & std  & 22.527      & 0.523       & 0.234      & 22.382  & 3.806       \\
                               & min  & 28.0        & 3.804       & 4.999      & 6.508   & 0.0         \\
                               & max  & 128.0       & 6.411       & 6.112      & 108.073 & 30.897      \\ \hline
\multirow{4}{*}{Top-$p$}       & mean & 86.706      & 5.163       & 5.64       & 41.218  & 4.002       \\
                               & std  & 21.166      & 0.511       & 0.223      & 23.321  & 3.727       \\
                               & min  & 25.0        & 3.795       & 4.952      & 4.783   & 0.0         \\
                               & max  & 130.0       & 6.33        & 6.191      & 105.246 & 30.897      \\ \hline
\end{tabular}
\end{table}

Table~\ref{tab:stats-text-dec-strategies} displays the lexical characteristics averaged over decoding strategies. There is little variability with respect to average text length, average word length, and Zipf frequency across decoding strategies (although beware that these metrics are computed after the post-processing of the generated texts, see Section~\ref{sec:post-processing-generated-texts}). The greatest differences occur with regard to the Flesch reading ease: greedy search and beam search appear to produce slightly more difficult texts than random sampling and top-$p$ sampling. Concerning surprisal, random sampling produces, on average, words with the highest surprisal value, although this stands somewhat in contrast to random sampling having the lowest Flesch reading ease score. 
Tables detailing these lexical characteristics for each model and each decoding strategy separately can be found in Appendix~\ref{appendix:descriptive-stats-stimuli}.

\begin{figure}[]
    \centering

    \begin{minipage}[b]{0.45\textwidth}
        \centering
        \includegraphics[width=\textwidth]{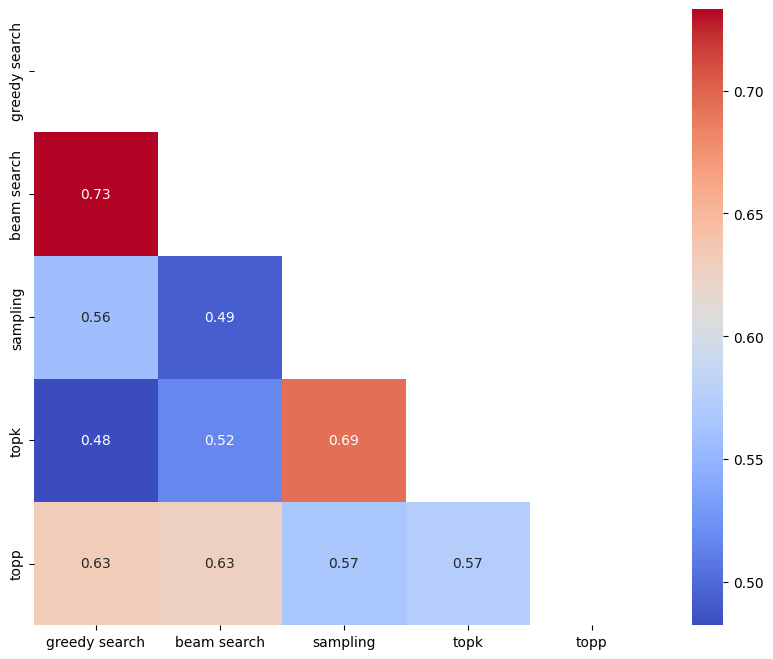}
        \caption{Text length.}
        \label{fig:subfig1}
    \end{minipage}
    \hfill
    \begin{minipage}[b]{0.45\textwidth}
        \centering
        \includegraphics[width=\textwidth]{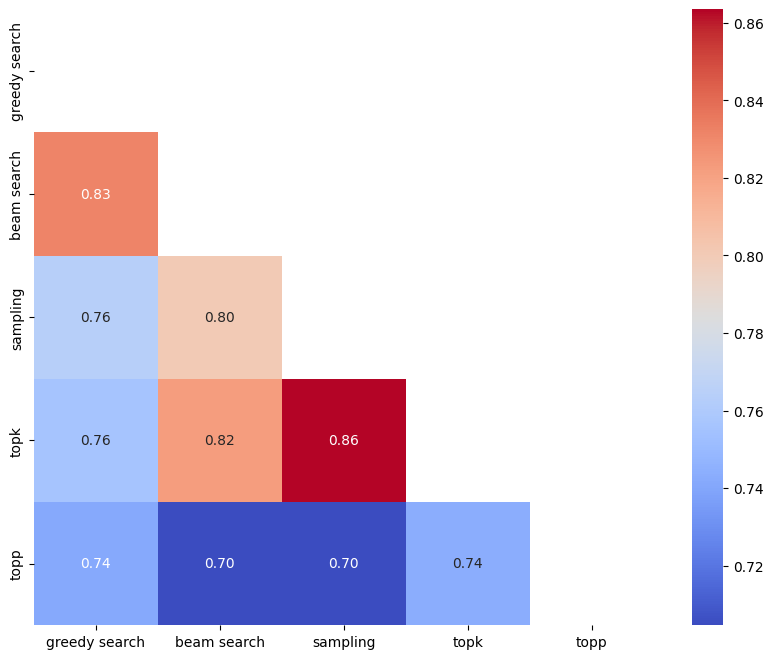}
        \caption{Word length.}
        \label{fig:subfig2}
    \end{minipage}
    
    \vspace{0.5cm} 
    \begin{minipage}[b]{0.45\textwidth}
        \centering
        \includegraphics[width=\textwidth]{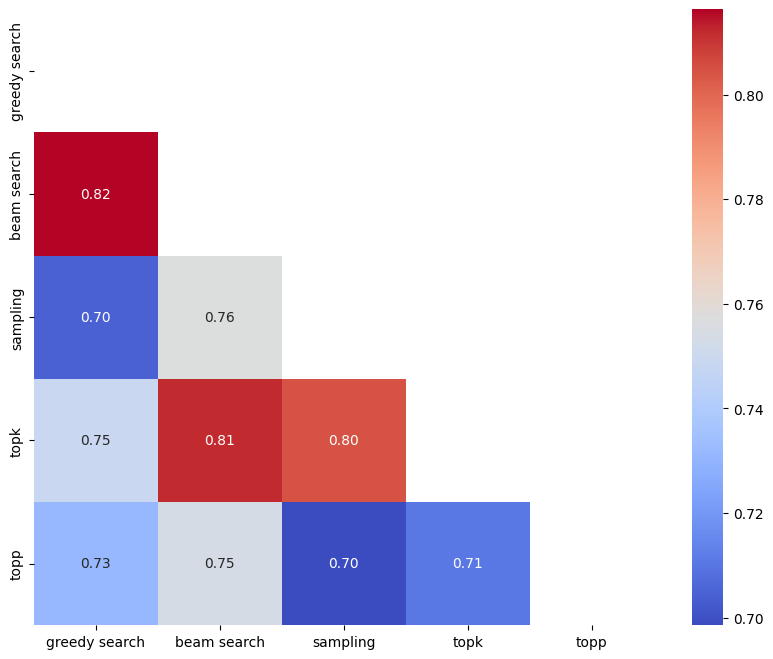}
        \caption{Zipf frequency.}
        \label{fig:subfig3}
    \end{minipage}
    \hfill
    \begin{minipage}[b]{0.45\textwidth}
        \centering
        \includegraphics[width=\textwidth]{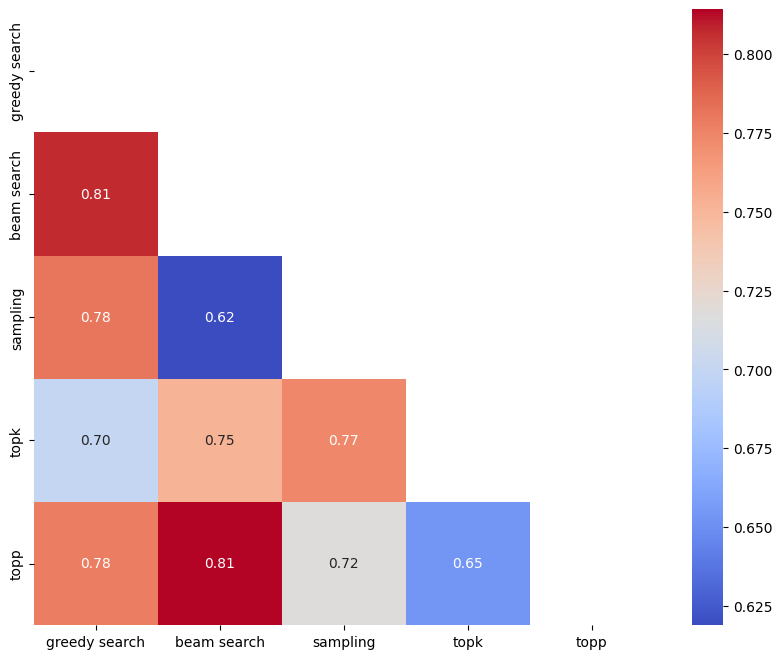}
        \caption{GPT-2 \textit{base} surprisal.}
        \label{fig:subfig4}
    \end{minipage}

\caption{Correlations between texts produced by different decoding strategies for four different text metrics.}
\label{fig:stats-correlations-dec-strategies}
\end{figure}

We further compute the correlations between texts generated with the same decoding strategy (across all three LLMs) for the lexical characteristics text length, word length, Zipf frequency, and GPT-2 \textit{base} surprisal, as depicted in Figure~\ref{fig:stats-correlations-dec-strategies}. The greatest differences in correlation strength between the different decoding strategies occurs when being computed for the text length feature, ranging from 0.48 (correlation between greedy search and top-$k$ sampling) to 0.73 (beam search and greedy search). The least variability across correlation statistics occurs regarding surprisal extracted from the smaller GPT-2 \textit{base} model, with the strongest correlations occurring for beam search and greedy search, and for beam search and top-$p$ sampling. Since the decoding strategies can be grouped into two categories (likelihood maximization methods, which include greedy search and beam search; and stochastic methods, which include random sampling, top-$k$ sampling, and top-$p$ sampling), one might expect that those belonging to the same category displayed stronger correlations as opposed to across categories. However, for all lexical metrics except surprisal, top-$p$ sampling seems to correlate less strongly with the other stochastic methods than with the likelihood maximization methods.

\section{Eye-Tracking Experiment}
\label{sec:eye-tracking-experiment}

\subsection{Participants}
\label{sec:participants}

In total, 119 native speakers of English and with normal or corrected-to-normal vision participated in the experiment in Zurich, Switzerland. Out of these, we discarded the data of 12 participants due to insufficient data quality or bad calibration, which results in data of a total of 107 participants in the corpus. The participants are from various backgrounds and were recruited in a variety of ways: via the University of Zurich; Facebook groups such as \textit{Expats in Switzerland}, \textit{English-speaking jobs in Zurich}, and \textit{Au-Pairs in Switzerland}; newsletters; the Democrats Abroad Switzerland; and the American Women's Club Switzerland. Before commencing the experiment, they filled out a short questionnaire that inquired about their age, gender, amount of sleep, handedness, visual aids, possible second native languages, their English variant, astigmatism, myopia and hyperopia, and neurological disorders. An overview of selected characteristics is depicted in Table~\ref{tab:participant-stats}. All participant characteristics and information can be found in the file \texttt{participant\_info/participant\_info.csv}. 

Participation requirements entailed being 18 years old or older, having normal or corrected-to-normal vision, and being a native English speaker. There was no restriction pertaining to the variety of English; the distribution of English varieties spoken by the participants is displayed in Table~\ref{tab:english-variants}. Participants were requested to not consume alcohol the night before or the day of the experiment. In case of wearing contact lenses, they were asked to wear their glasses, if available, and bring the contact lenses along. They were further asked not to wear eye makeup. Participants received a compensation of 30 CHF for an estimated 1 hour of participation.

\begin{table}[]
\caption{Overview of the participants' mean age and a selection of other statistics.}
\label{tab:participant-stats}
\begin{tabular}{ccccccc}
\hline
\multirow{2}{*}{\shortstack{Number of \\ participants}} & \multirow{2}{*}{Gender} & \multirow{2}{*}{\shortstack{Mean age \\ $\pm$ sd}} & \multirow{2}{*}{\shortstack{Mean hours of \\ sleep $\pm$ (sd)}} & \multirow{2}{*}{\shortstack{Visual \\ aid\footnotemark[1]}} & \multirow{2}{*}{\shortstack{Myopia/ \\ hyperopia\footnotemark[2]}} & \multirow{2}{*}{Astigmatism} \\ 
 &  &  &  &  &  &  \\ \hline
\multirow{2}{*}{67} & \multirow{2}{*}{female} & \multirow{2}{*}{32.5 $\pm$ 12.0} & \multirow{2}{*}{7.3 $\pm$ 1.0} & g: 36; s: 2 & m: 31; h: 7 & yes: 15 \\
 &  &  &  & n: 29 & n: 29 & no: 52 \\ \hline
\multirow{2}{*}{39} & \multirow{2}{*}{male} & \multirow{2}{*}{37.1 $\pm$ 15.7} & \multirow{2}{*}{7.0 $\pm$ 1.0} & g: 21; s: 2 & m: 15; h: 8 & yes: 8 \\
 &  &  &  & n: 16 & n: 16 & no: 31 \\ \hline
\multirow{2}{*}{1} & \multirow{2}{*}{other} & \multirow{2}{*}{33} & \multirow{2}{*}{5} & g: 1; n: - & m: 1; h: - & yes: 1 \\ 
 &  &  &  &  n: - & n: -  & no: -  \\ \hline
\multirow{2}{*}{107} &  & \multirow{2}{*}{34.1 $\pm$ 13.5} & \multirow{2}{*}{7.2 $\pm$ 1.0} & g: 53; s: 4 & m: 47; h: 15 & yes: 24 \\
 &  &  &  & n: 50 & n: 45 & no: 83 \\ \hline
\end{tabular}
\footnotetext[1]{``g'' $=$ glasses; ``s'' $=$ soft contact lenses; ``n'' $=$ no visual aid}
\footnotetext[2]{``m'' $=$ myopic (short-sighted); ``h'' $=$ hyperopic (far-sighted), ``n'' $=$ neither}
\end{table}

\begin{table}[]
\caption{Distribution of variants of English across participants.}
\label{tab:english-variants}
\begin{tabular}{lc}
\hline
English Variety & Number of Speakers \\ \hline
American      & 45              \\
British       & 33              \\
Canadian      & 8               \\
Australian    & 7               \\
Indian        & 3               \\
Irish         & 3               \\
Scottish      & 3               \\
New Zealand   & 2               \\
Gibraltar     & 1               \\
South African & 1               \\
Zimbabwe      & 1               \\ \hline
\end{tabular}
\end{table}

\subsection{Experiment Design}
\label{sec:experiment-design}

The 42 items in their 14 experimental conditions, where each condition consists of a combination of model and decoding strategy, were distributed across 14 lists in a Latin Square design to ensure that each participant sees each item only once, and each condition is presented to a participant the same number of times. Upon presentation, the order of the experimental items in each list was randomized. Each of the 42 trials consisted of four distinct parts: a header, the stimulus text, a comprehension question, and two rating questions. 
Each part was presented on its own screen. 
All text was presented in a mono-spaced font (Courier, font size 14) with double line spacing.

\paragraph{Header}
The header is a sentence that introduces the subsequent text. Since the different texts are quite varied not only in content but also in form, the header's purpose is to avoid the confusion of the reader and an initial skimming of the new text at hand. The headers are modeled as closely after the original model prompts as possible without revealing their original nature, \emph{i.e.}, that they were in fact prompts for a language model. We opted against displaying the prompts in their original form lest participants might discern that the texts were machine-generated. Furthermore, some prompts, such as the ones used for the summarization tasks, were too long to be presented on one single screen and would have anticipated the subsequent text's content in a way that would have biased the readers' eye movements. 

\paragraph{Stimulus}
The header is followed by the post-processed and validated generated text. 
Otherwise the sequences are displayed as predicted by the models, including line breaks and white-space characters. However, for some texts, mostly the dialogues and some news articles, the models consistently predicted double newlines. In these cases, we removed one of the newline characters to ensure that the texts fit onto the presentation screen. This does not affect analyses with the data, however, as newlines are not areas of interest on their own, thus no fixations or reading measures are mapped onto them. From a psycholinguistic point of view, line breaks are important because words immediately preceding or following a line break exhibit a fixation pattern that would be different than if their position was further away from a line break. From a model perspective, these newline characters are also important, as tokens immediately following a newline character are conditioned on that newline character's probability distribution and not on the distribution of the word preceding the newline. That being said, the texts as presented in the experiments exhibit more line breaks than originally predicted by the models, as they would not have fitted onto the screen otherwise. Assuming the psycholinguistic perspective, this does not change anything about the interpretation of fixations landing on tokens before or after newlines, though from the model perspective, especially when taking into account the transition scores of the generated tokens, there might be a slight bias, which was unavoidable unfortunately.

\paragraph{Comprehension Question}
Each text is followed by a multiple-choice comprehension question with four possible answers, out of which one answer is true. They serve the purpose of ensuring participants remain attentive and would allow to exclude data where people clicked through the texts and questions inattentively. The comprehension questions require inference from the text that was read, but with their main purpose being to maintain participants' attention levels, they do not attempt to quantify their text comprehension. 

\paragraph{Rating Questions}
After the comprehension question, participants answered two rating questions: the first inquired how subjectively difficult they deemed the text, and the second how engaging they found the text. Both questions were answered on a scale from 1 to 5.

\subsection{Procedure}
\label{sec:procedure}

The duration of one experiment session was around 60 minutes. Upon arrival, participants were asked to sign a declaration of consent and fill out the demographic questionnaire. Subsequently, the experimenter outlined the procedure of the experiment and provided them with instructions, which were both delivered verbally by the experimenter and again on the screen at the onset of the experiment: they were told to read the texts naturally, the way they would read a book or a newspaper, and that there were no time constraints; they should remain as still as possible and avoid asking questions or moving while reading the stimulus text. Upon calibration, the experiment started with a practice trial which would be discarded from the data. Before the appearance of both the header and the text on the presentation screen, participants were instructed to fixate on a fixation point at the position where the subsequent header and text would start. Drift correction was only applied if absolutely necessary. Participants could move from one screen to the next at their own pace by pressing the \textit{Space} bar. 
Responses to the questions were given with the numbers 1 to 4 (comprehension questions) or 1 to 5 (rating questions) on the keyboard. After reading the header or the text, participants were asked to focus on a blue sticker attached to the bottom right corner of screen before pressing the \textit{Space} bar to proceed; this ensured that they did not fixate on the text at random between their finishing their reading process and pressing the key to continue. The experiment was divided into three sections; there was a five-minute break after 14 and after 28 experimental stimuli that allowed participants to quickly rest their eyes and get up. Re-calibration was performed after each break, and also in-between if necessary.

\subsection{Data Acquisition and Setup}
\label{sec:data-acquisition-and-setup}

The data acquisition took place in a controlled setting. 
Participants were placed in a sound-insulated and windowless room with controlled lighting. Participants put their head on a head-and-chin rest to limit head movement. They were instructed to remain immobile during the experiment, and were encouraged to rest their eyes and move their body during the breaks. They were seated at a height-adjustable table to warrant a constant eye-to-screen distance across participants. The eye-to-screen distance was 60\,cm, with a 24\,inch screen, and the distance from the eyes to the eye-tracker was 55\,cm. Owing to the height-adjustable table, the head-and-chin rest did not need to be adjusted. We employed a resolution of 1280$\times$1024\,pixels, which subsets the original screen size to one of 31\,cm height and 43\,cm width. The monitor refresh rate was 60\,Hz. The experiment was implemented with Experiment Builder~\citep{SRResearch2020}.

Before commencing the experiment, the experimenter determined the participant's dominant eye, which was then used for monocular eye-tracking. We tracked the eye position with a video-based infrared eye-tracker (EyeLink Portable Duo\footnote{\url{https://www.sr-research.com/eyelink-portable-duo/}}), at a sampling rate of 2000 Hz. The eye-tracker was manually calibrated with a 9-point grid, the fixation points of which were displayed in random order. In a second step, the validation ensured that the error between two measurements at any point was less than 0.5$^\circ$. 

\section{Eye-Tracking Data: Pre-Processing}
\label{sec:preprocessing-eye-tracking-data}

\subsection{Pre-Processing of Raw Data}
\label{sec:raw-data}

The data files written by the eye-tracker are non-human readable \texttt{edf} files. We converted them to \texttt{asc} with the \texttt{edf2asc} tool provided by SR Research. These \texttt{asc} files contain the data for one session, which not only include the recorded eye movement coordinates but also all metadata, such as button presses, drift corrections, re-calibrations, answers to the comprehension questions, and messages that indicate the onset and offset of individual trials and screens. These \texttt{asc} files are parsed and converted to \texttt{csv} to extract the relevant samples for each trial. The \texttt{csv} files consist of one line per sample, \emph{i.e.}, 2000 lines per second, which include timestamps as well as the  x- and y-coordinates of where the tracked eye focussed. We tracked each participant's dominant eye, so whether we tracked the left or the right one was participant-dependent, though all \texttt{csv} files bear the column names \texttt{x} and \texttt{y} for simplification purposes. The information on which eye was tracked for each participant can be found in \texttt{participant\_info/participant\_info.csv}.

\subsection{Fixation Extraction}
\label{sec:fixation-extraction}

\begin{figure}
    \centering

    \begin{subfigure}{1\textwidth}
    \centering
    \includegraphics[width=1\textwidth]{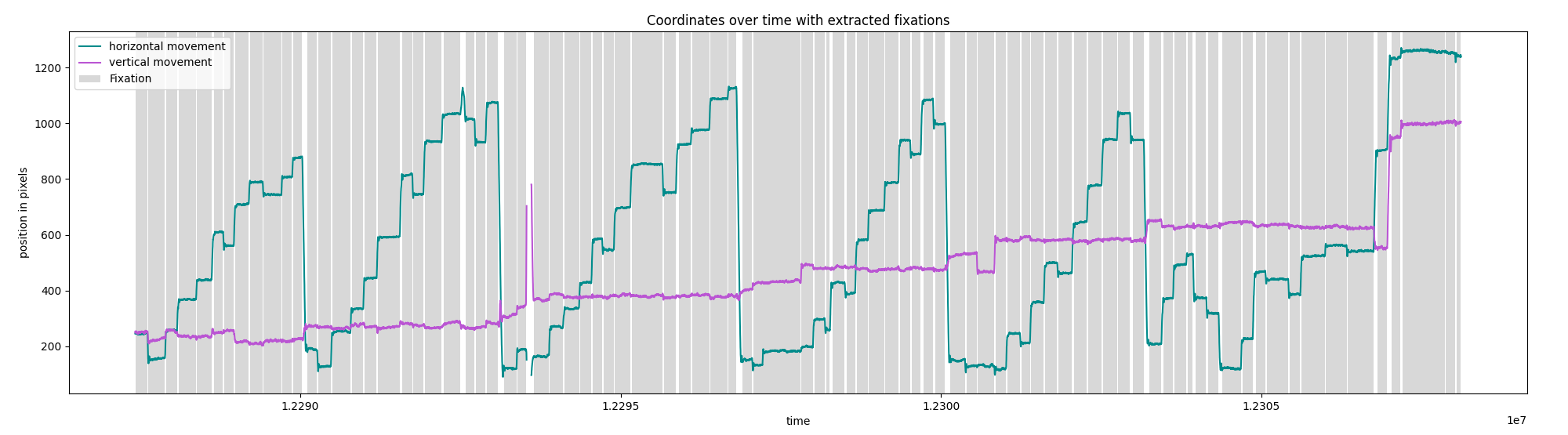}
    \caption{Subject 23 reading item 11.}
    \label{fig:et23-fixations-coordinates}
    \end{subfigure}

    \begin{subfigure}{1\textwidth}
    \centering
    \includegraphics[width=1\textwidth]{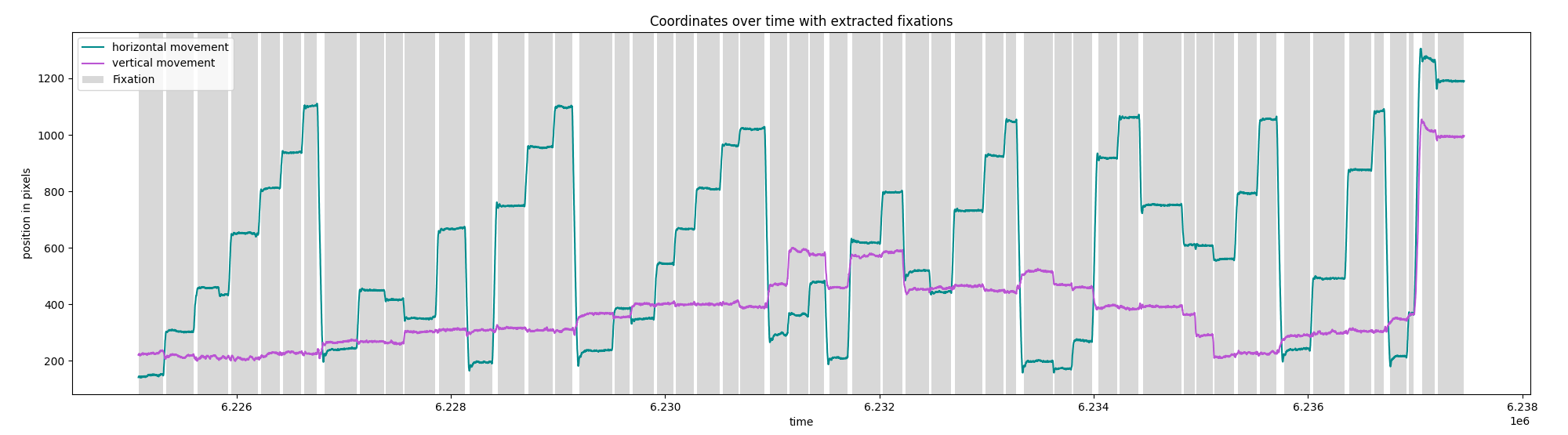}
    \caption{Subject 109 reading item 11.}
    \label{fig:et109-fixations-coordinates}
    \end{subfigure}

    \caption{Horizontal and vertical eye movements during reading of one text, with the grey bars indicating the fixations detected by the algorithm.}
    \label{fig:fixations-coordinates}
\end{figure}

In order to group the raw data samples into fixations and saccades, we used the microsaccade detection algorithm introduced by \citet{engbert2003microsaccades}. Saccades can be identified via their velocities, \emph{i.e.}, they are outliers in velocity space. Thus the time series of coordinates of eye positions was first transformed into velocities as a weighted moving average over five data samples to suppress noise~\citep{engbert2006microsaccades}, and then the detection threshold is computed as a multiple of the standard deviation of the velocity distribution. The detection thresholds are computed separately not only for each subject, but for each individual text, as proposed by~\citet{engbert2015microsaccade}, relative to the noise level, and they are multiplied by the threshold factor. Since the detection threshold is chosen with respect to the noise level of a single text, the algorithm is robust regarding different noise levels between texts as well as across participants. There are different methods to compute the threshold; we employed the one proposed by~\citet{engbert2015microsaccade}. Figure~\ref{fig:fixations-coordinates} shows the eye movements in terms of coordinates over time for two participants in two trials, where the green graph depicts the horizontal eye movement and the pink graph the vertical eye movement. The saccades detected by the algorithm are the white bars, and the grey bars are the fixations consequently. Although the two participants exhibit quite different reading behavior, the microsaccade detection algorithm efficiently and correctly groups and detects fixations and saccades for both of them.

\begin{figure}[h]
    \centering
    \begin{subfigure}{0.48\textwidth}
        \centering
        \includegraphics[width=\textwidth]{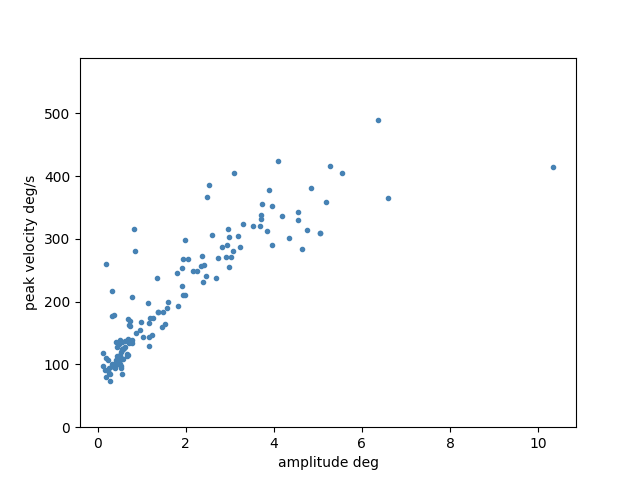}
        \caption{Subject 23 reading item 11.}
        \label{fig:et23-saccade-ampl-vel}
    \end{subfigure}
    \hspace{0.02\textwidth}
    \begin{subfigure}{0.48\textwidth}
        \centering
        \includegraphics[width=\textwidth]{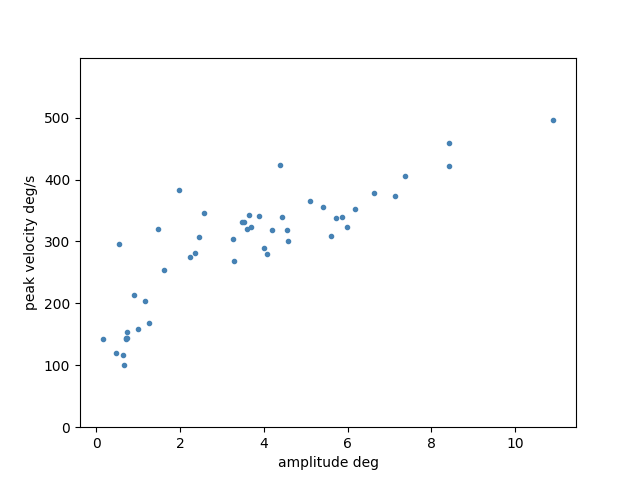}
        \caption{Subject 109 reading item 11.}
        \label{fig:et109-saccade-ampl-vel}
    \end{subfigure}
    \caption{Saccade peak velocity in degrees of visual angle per second plotted over saccade amplitude in degrees.}
    \label{fig:saccades-ampl-vel}
\end{figure}

Since microsaccades, as well as saccades, are ballistic movements, \emph{i.e.}, rapid and forceful motions that leave little to no time for feedback mechanisms to make adjustments during the movement, they exhibit a fixed relation between peak velocity and amplitude. An inspection of this linear relationship between peak velocity and amplitude can thus serve as a kind of sanity check of the validity of the algorithm, \emph{i.e.}, whether or not the thresholds are correctly estimated and the saccades are accurately detected. Figure~\ref{fig:saccades-ampl-vel} depicts the peak velocities of saccades plotted over their amplitudes for two subjects reading two trials. The linear relationship is clearly discernible. Moreover, we also see that the algorithm sets an accurate upper bound regarding peak velocities by removing peak velocities greater than 500 degrees of visual angle per second. While it is indeed possible for human eyes to reach peak saccade velocities up to 1000 degrees per second, the upper bound for saccades during reading is around 500 degrees per second~\citep{holmqvist2011eye}; measurements with greater velocities are results of artefacts or blinks. Pertaining to saccade amplitude, there is no upper bound that can be systematically set. Figure~\ref{fig:saccades-ampl-vel} does indeed display outliers regarding the amplitude for both subjects where they moved their eyes entirely away from the text, but these outliers are removed in the manual fixation correction stage (see Section~\ref{sec:manual-fixation-correction}).

\subsection{Manual Fixation Correction}
\label{sec:manual-fixation-correction}

\begin{figure}
    \centering
    \includegraphics[width=0.9\textwidth]{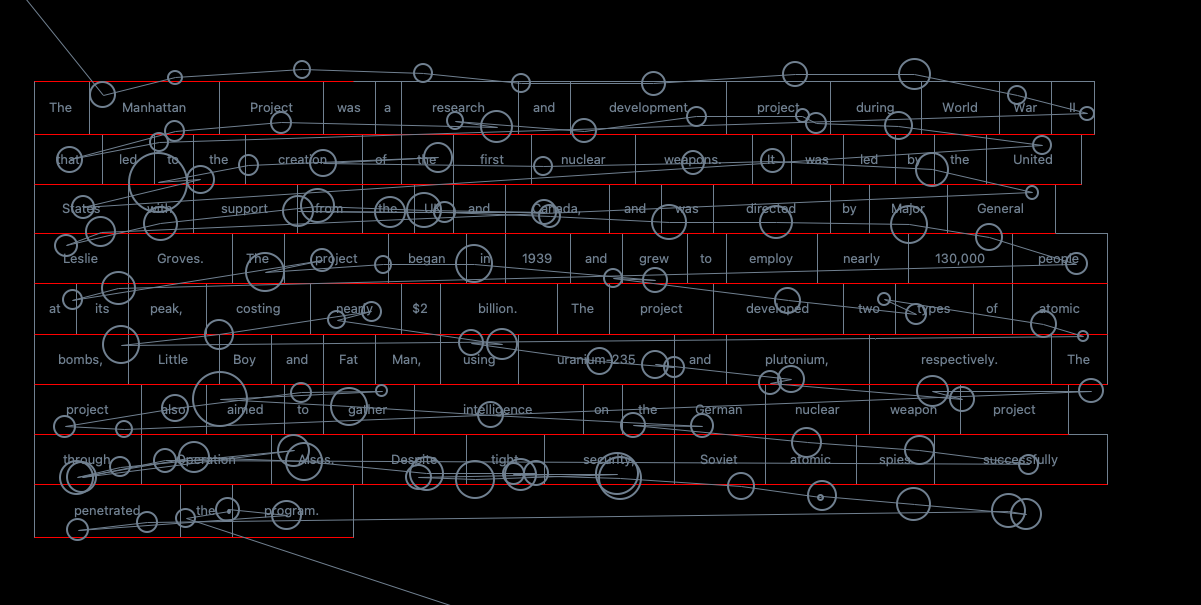}
    \caption{An example of vertical drift during the eye-tracking experiment. Subject 44 reading item43.}
    \label{fig:vertical-drift-example}
\end{figure}

A common issue that arises when conducting eye-tracking experiments is the so-called vertical drift, which means that there occurs a gradual displacement of the recorded coordinates over time~\citep{carr2022algorithms}. Phrased differently, coordinates are being recorded as being above or below the line that the participant is actually reading, which in turn results in fixations being placed above or below the lines on which they actually landed. This is particularly precarious when it comes to eye-tracking on the passage level, where participants read multi-line texts. 
In such settings, vertical drift results in fixations being mapped to interest areas on the line above or below, which can distort the data. Vertical drift can occur even if the initial calibration was good and it is likely the result of deteriorating calibration over time or in the corners of the screen, or of subtle head movements, pupil dilations, teary eyes, and artifacts. Moreover, this vertical drift is usually not systematic across the entire screen and thus cannot be eliminated by drift correction. Figure~\ref{fig:vertical-drift-example} illustrates an example of a subject exhibiting vertical drift: there is an upward drift towards the middle of the screen, and a downward drift at the end of the line towards the bottom of the screen. These kind of patterns cannot be remedied by drift correction. More examples of vertical drifts can be found in Appendix~\ref{appendix:manual-fixation-correction}.

In addition to vertical drift, there can also be instances of horizontal drift, which is unfortunately less recognizable than the vertical drift. However, since the areas of interest consist of words, we strongly assume that a possible horizontal measurement error will lead to only very few wrong fixation-to-word mappings, as it would have to be greater than the distance between the recorded fixation and the word boundary of its true area of interest.

Another problem leading to fixations being mapped to the wrong areas of interest is unrelated to the occurrence of vertical drift: participants might not immediately start reading once the text appears, but many first jump to the middle or the end of the text. Although these initial fixations might be considered part of a subject's reading behavior, as they orient themselves within the text layout, they influence and distort the computation of reading measures. For instance, if a subject first jumps to the middle of the screen and then back to the beginning of the text, all words up to that first fixation in the middle would be attributed a first-pass reading time of zero. 

Furthermore, some participants also move their eyes out of the text during reading as well. These are also fixations that are mapped to the closest area of interest, though they in fact do not belong to the true reading behavior. Moreover, since we have asked participants to move their eyes onto a sticker at the bottom right corner of the screen upon finishing their reading, this bottom right area also contains several fixations. However, this kind of eye movement behavior that is not inherent to the actual text reading is very easily discernible.

To amend the above-mentioned problems, we therefore performed a manual fixation correction, during which we mapped all fixations exhibiting a vertical displacement to their 
nearest vertical areas of interest and removed the fixations that resulted from scanning the text before reading and jumping out of the text while reading.  Although human eye movements in reading are interspersed with regressions and forward saccades, the underlying sequential nature of reading is still very well visible. While there might always be a chance of mapping errors, it is usually easy for a human annotator experienced with eye movements in reading to recognize the vertical drift and perform the fixation correction, a practice that is common in many eye-tracking labs.


\subsection{Computation of Reading Measures}
\label{sec:reading-measures-computation}

From the fixation data, we computed a range of reading measures commonly used in reading research. The interest areas for the reading measures are equal to the ones of the fixation data; each reading measure is mapped to a white-space delimited word. This also entails that punctuation marks are not areas of interest on their own, but fixations landed on them are mapped to the preceding word, or to the subsequent word in case of opening quotation marks or parentheses. The abbreviations and definitions of the computed reading measures are provided in Table~\ref{tab:reading-measures-definition}.

\begin{table}[htbp]
\caption{Definition and abbreviation of reading measures computed for EMTeC.}
\label{tab:reading-measures-definition}
\centering
\footnotesize
\begin{tabularx}{\textwidth}{>{\raggedright\arraybackslash}p{2.8cm} l X}
\hline
\textbf{Measure} & \textbf{Abbr.} & \textbf{Definition} \\
\hline
\multicolumn{3}{l}{\textit{Continuous measures in ms}} \\
\hline
first-fixation duration & FFD & duration of the first fixation on a word if this word is fixated in first-pass reading, otherwise 0 \\
first duration & FD & duration of the first fixation on a word (identical to FFD if not skipped in the first-pass) \\
first-pass reading time & FPRT & sum of the durations of all first-pass fixations on a word (0 if the word was skipped in the first-pass) \\
single-fixation duration & SFD & duration of the only first-pass fixation on a word, 0 if the word was skipped or more than one fixations occurred in the first-pass (equals FFD in case of a single first-pass fixation) \\
first-reading time & FRT & sum of the duration of all fixations from first fixating the word (independent if the first fixations occurs in first-pass reading) until leaving the word for the first time (equals FPRT in case the word was fixated in the first-pass) \\
total-fixation time & TFT & sum of all fixations on a word (FPRT+RRT) \\
re-reading time & RRT & sum of the durations of all fixations on a word that do not belong to the first-pass (TFT-FPRT) \\
inclusive regression-path duration & RPD\_inc & sum of all fixation durations starting from the first first-pass fixation on a word until fixation a word to the right of this word (including all regressive fixations on previous words), 0 if the word was not fixated in the first-pass (RPD\_exc+RBRT) \\
exclusive regression-path duration & RPD\_exc & sum of all fixation durations after initiating a first-pass regression from a word until fixating a word to the right of this word, without counting fixations on the word itself (RPD\_inc-RBRT) \\
right-bounded reading time & RBRT & sum of all fixation durations on a word until a word to the right of this word is fixated (RPD\_inc-RPD\_exc) \\
\hline
\multicolumn{3}{l}{\textit{Binary measures}} \\
\hline
fixation & Fix & 1 if the word was fixated, otherwise 0 (FPF or RR) \\
first-pass fixation & FPF & 1 if the word was fixated in the first-pass, otherwise 0 \\
first-pass regression & FPReg & 1 if a regression was initiated in the first-pass reading of the word, otherwise 0 (RPD\_exc) \\
re-reading & RR & 1 if the word was fixated after the first-pass reading, otherwise 0 (RRT) \\
\hline
\multicolumn{3}{l}{\textit{Discrete measures}} \\
\hline
total fixation count & TFC & total number of fixations on that \\
incoming saccade length & SL\_in & length of the saccade that leads to first fixation on a word in number of words; positive sign if the saccade is a progressive one, negative sign if it is a regression \\
outgoing saccade length & SL\_out & length of the first saccade that leaves the word in number of words; positive sign if the saccade is a progressive one, negative sign if it is a regression; 0 if the word is never fixated \\
total count of incoming regressions & TRC\_in & total number of regressive saccades initiated from this word \\
total count of outgoing regressions & TRC\_out & total number of regressive saccades landing on this word \\
\hline
\end{tabularx}
\end{table}

\section{Analyses}

\subsection{Descriptive Statistics of the Reading Measures}
\label{sec:descriptive-statistics}

We provide descriptive statistics of a selection of reading measures: the continuous measures \textit{first-fixation duration} (FFD; duration of the first fixation on a word if it was fixated in first-pass reading), \textit{first-pass reading time} (FPRT; sum of the durations of all first-pass fixations on a word), \textit{total-fixation time} (TFT; sum of the durations of all fixations on a word), and \textit{re-reading time} (RRT; sum of the durations of all fixations on a word that do not belong to the first-pass); and the binary measures \textit{fixation} (Fix; whether or not a word was fixated) and \textit{re-reading} (RR; whether or not a word was fixated after the first-pass reading).

Table~\ref{tab:stats-rms-models} depicts statistics on the reading measures with respect to outputs of the individual models. We observe minimal variation both with respect to mean values as well as standard deviations between models, indicating that the reading behavior does not drastically change according to which model the respective texts were generated with. WizardLM seems to have slightly higher values across all metrics; for instance, words in texts generated by WizardLM display a higher re-reading time than words of texts originating from the other two models.

\begin{table}[htb!]
\caption{Descriptive statistics of a selection of reading measures for each model. See Table~\ref{tab:reading-measures-definition} for a definition of the different reading measures.\protect\footnotemark[1]}
\label{tab:stats-rms-models}
\begin{tabular}{llll}
\hline
& Phi-2 & Mistral & WizardLM \\ \hline
\multicolumn{4}{c}{\textit{Mean and standard deviation of continuous measures in ms}}   \\ \hline
FFD & 223.165 $\pm$ 101.362 & 222.166 $\pm$ 103.895 & 222.414 $\pm$ 100.985 \\
FPRT & 268.674 $\pm$ 154.69 & 265.261 $\pm$ 151.312 & 266.831 $\pm$ 151.84 \\
TFT & 326.477 $\pm$ 226.091 & 325.399 $\pm$ 230.474 & 327.748 $\pm$ 231.686 \\
RRT & 295.616 $\pm$ 213.118 & 300.596 $\pm$ 224.297 & 304.003 $\pm$ 229.13 \\ \hline
\multicolumn{4}{c}{\textit{Mean proportions of the binary measures}}       \\ \hline
Fix & 0.694  & 0.694  & 0.695 \\
RR & 0.221  & 0.22  & 0.221 \\ \hline
\end{tabular}
\footnotetext[1]{For the statistics of FFD, FPRT, TFT, and RRT, words with a value of 0 (not fixated) were excluded.}
\end{table}

Statistics on the same reading measures pertaining to outputs generated with different decoding strategies are provided in Table~\ref{tab:stats-rms-dec-strategies}. Again, there is little variability between the different decoding strategies regarding \textit{first-fixation duration} and \textit{first-pass reading time}. Concerning \textit{total-fixation time}, it appears that words in texts generated with beam search are fixated for a longer time overall, while those generated with greedy search have the overall lowest total fixation durations. A similar pattern can be observed with respect to \textit{re-reading time}: words in the beam search outputs that are re-read are fixated for a longer time than words of other outputs; and greedy search output words are re-read for the shortest amount of time. This higher value of TFT in poems is probably accounted for by a longer re-reading time compared to the other text types.

\begin{table}[htb!]
\centering
\caption{Descriptive statistics of a selection of reading measures for each decoding strategy. See Table~\ref{tab:reading-measures-definition} for a definition of the different reading measures.\protect\footnotemark[1]}
\label{tab:stats-rms-dec-strategies}
\begin{tabular}{llll}
\hline
           & Greedy search           & Beam search             &              \\ \hline
\multicolumn{3}{c}{\textit{Mean and standard deviation of continuous measures in ms}} \\ \hline
FFD        & 221.162 $\pm$ 101.095   & 223.534 $\pm$ 106.072  &              \\
FPRT       & 265.048 $\pm$ 152.407   & 266.682 $\pm$ 153.026  &              \\
TFT        & 323.098 $\pm$ 228.944   & 331.051 $\pm$ 239.462  &              \\
RRT        & 296.862 $\pm$ 222.448   & 309.743 $\pm$ 237.347  &              \\ \hline
\multicolumn{3}{c}{\textit{Mean proportions of the binary measures}} \\ \hline
Fix        & 0.693     & 0.692  &              \\
RR         & 0.22        & 0.223    &              \\ \hline
\end{tabular}
\quad
\begin{tabular}{llll}
\hline
           & Sampling                & Top-$k$                 & Top-$p$        \\ \hline
\multicolumn{4}{c}{\textit{Mean and standard deviation of continuous measures in ms}} \\ \hline
FFD        & 223.013 $\pm$ 102.286   & 223.297 $\pm$ 102.994   & 221.962 $\pm$ 99.678  \\
FPRT       & 267.644 $\pm$ 153.084   & 267.573 $\pm$ 151.439   & 266.781 $\pm$ 152.463  \\
TFT        & 329.343 $\pm$ 233.574   & 325.505 $\pm$ 224.761   & 325.131 $\pm$ 224.759  \\
RRT        & 304.87 $\pm$ 228.032    & 296.272 $\pm$ 214.337   & 297.34 $\pm$ 216.7     \\ \hline
\multicolumn{4}{c}{\textit{Mean proportions of the binary measures}} \\ \hline
Fix        & 0.697       & 0.696   & 0.693  \\
RR         & 0.222       & 0.22    & 0.218 \\ \hline
\end{tabular}
\footnotetext[1]{For the statistics of FFD, FPRT, TFT, and RRT, words with a value of 0 (not fixated) were excluded.}
\end{table}

More distinct differences emerge when examining the reading measures with respect to the text types, as displayed in Table~\ref{tab:stats-rms-text-types}. While \textit{fiction} has the lowest \textit{first-fixation duration} and \textit{first-pass reading time}, \textit{poetry} has by far the longest \textit{first-fixation durations} and \textit{first-pass reading times}. This is also reflected in \textit{total-fixation time}: the words in the poems are, on average, fixated for the longest amount of time, while those in fictional texts are fixated on for the shortest duration.

\begin{table}[htb!]
\centering
\caption{Descriptive statistics of a selection of reading measures for each text type. See Table~\ref{tab:reading-measures-definition} for a definition of the different reading measures.\protect\footnotemark[1]}
\label{tab:stats-rms-text-types}
\begin{tabular}{llll}
\hline
& Non-fiction & Fiction & Poetry \\ \hline
\multicolumn{4}{c}{\textit{Mean and standard deviation of continuous measures in ms}} \\ \hline
FFD & 220.79 $\pm$ 99.935 & 211.246 $\pm$ 90.626 & 242.242 $\pm$ 120.157 \\
FPRT & 262.891 $\pm$ 145.72 & 239.13 $\pm$ 122.998 & 296.138 $\pm$ 171.722 \\
TFT & 309.957 $\pm$ 208.142 & 272.512 $\pm$ 169.158 & 404.261 $\pm$ 284.231 \\
RRT & 285.469 $\pm$ 207.77 & 253.637 $\pm$ 165.551 & 349.679 $\pm$ 262.204 \\ \hline
\multicolumn{4}{c}{\textit{Mean proportions of the binary measures}} \\ \hline
Fix & 0.679  & 0.645  & 0.736 \\
RR & 0.189 & 0.17 & 0.333 \\ \hline
\end{tabular}

\begin{tabular}{llll}

& Summarization & Synopsis & Key words \\ \hline
\multicolumn{4}{c}{\textit{Mean and standard deviation of continuous measures in ms}} \\ \hline
FFD & 223.367 $\pm$ 100.731 & 223.991 $\pm$ 105.15 & 225.396 $\pm$ 105.051 \\
FPRT & 270.289 $\pm$ 154.528 & 273.739 $\pm$ 162.574 & 275.374 $\pm$ 162.256 \\
TFT & 341.677 $\pm$ 239.886 & 342.11 $\pm$ 246.012 & 338.416 $\pm$ 244.337 \\
RRT & 308.318 $\pm$ 226.802 & 310.199 $\pm$ 230.016 & 308.105 $\pm$ 236.737 \\ \hline
\multicolumn{4}{c}{\textit{Mean proportions of the binary measures}} \\ \hline
Fix & 0.724  & 0.718 & 0.696 \\
RR & 0.257  & 0.24  & 0.222 \\ \hline
\end{tabular}
\footnotetext[1]{For the statistics of FFD, FPRT, TFT, and RRT, words with a value of 0 (not fixated) were excluded.}
\end{table}

\subsection{Psycholinguistic Analysis}
\label{sec:psycholinguistic-analyses}

As a proof-of-concept, we present a series of analyses that assess the effects of various word-level psycholinguistic features on four reading measures: \textit{first-pass reading time} (FPRT; sum of the durations of all first-pass fixatinos on a word), \textit{total fixation time} (TFT; sum of the durations of all fixations on a word), \textit{fixation proportion} (Fix; whether or not a word was fixated), and \textit{first-pass regressions} (FPReg; whether a regression was initiated in the first-pass reading of a word). To do so, we deploy hierarchical linear-mixed models with reading measures $y_{ij}$ obtained for word $j\in\{1\dots J\}$, where $J$ is the total number of words across all stimulus texts\footnote{Note that we enumerate the words across all stimulus texts such that each word of each stimulus text has its own index.} read by participant $i$ as response variables, and normalized (z-score transformed) predictors \texttt{word length} $l_j$ (number of characters including punctuation), \texttt{Zipf frequency} $f_j$, \texttt{lexicalized surprisal} (GPT-2 \textit{base}) $s_j$, and \texttt{last in line} $t_j$ (binary: 1 if the word included a line-break), formalized in Equation~\ref{eq:lmm}
\begin{equation}\label{eq:lmm}
    y_{ij} = L(\beta_0 + \beta_{0i} + \beta_{1}{l}_{j} + \beta_{2}{f}_{j} + \beta_{3}{s}_{j}  +\beta_{4} t_j )\quad ,
\end{equation}
where $\beta_0$ denotes the global, and $\beta_{0i}$ the random intercept for subject $i$. $L(\cdot)$ denotes the linking function, in the case of the binary measures (\textit{first-pass regression}, \textit{fixation proportion}), $L(z)=\ln \frac{z}{1-z}$, with $y_{ij}$ following a Bernoulli distribution; and in case of the  for the continuous measures (\textit{first-pass reading time}, \textit{total fixation time}) $L(z)=\log z$, with $y_{ij}$ following a log-normal distribution. We run each model for $3000$ iterations including $1000$ warm-up iterations. We use standard uninformed priors, reported in Appendix~\ref{appendix:priors}.

\begin{figure}
    \centering
    \includegraphics[width=1\textwidth]{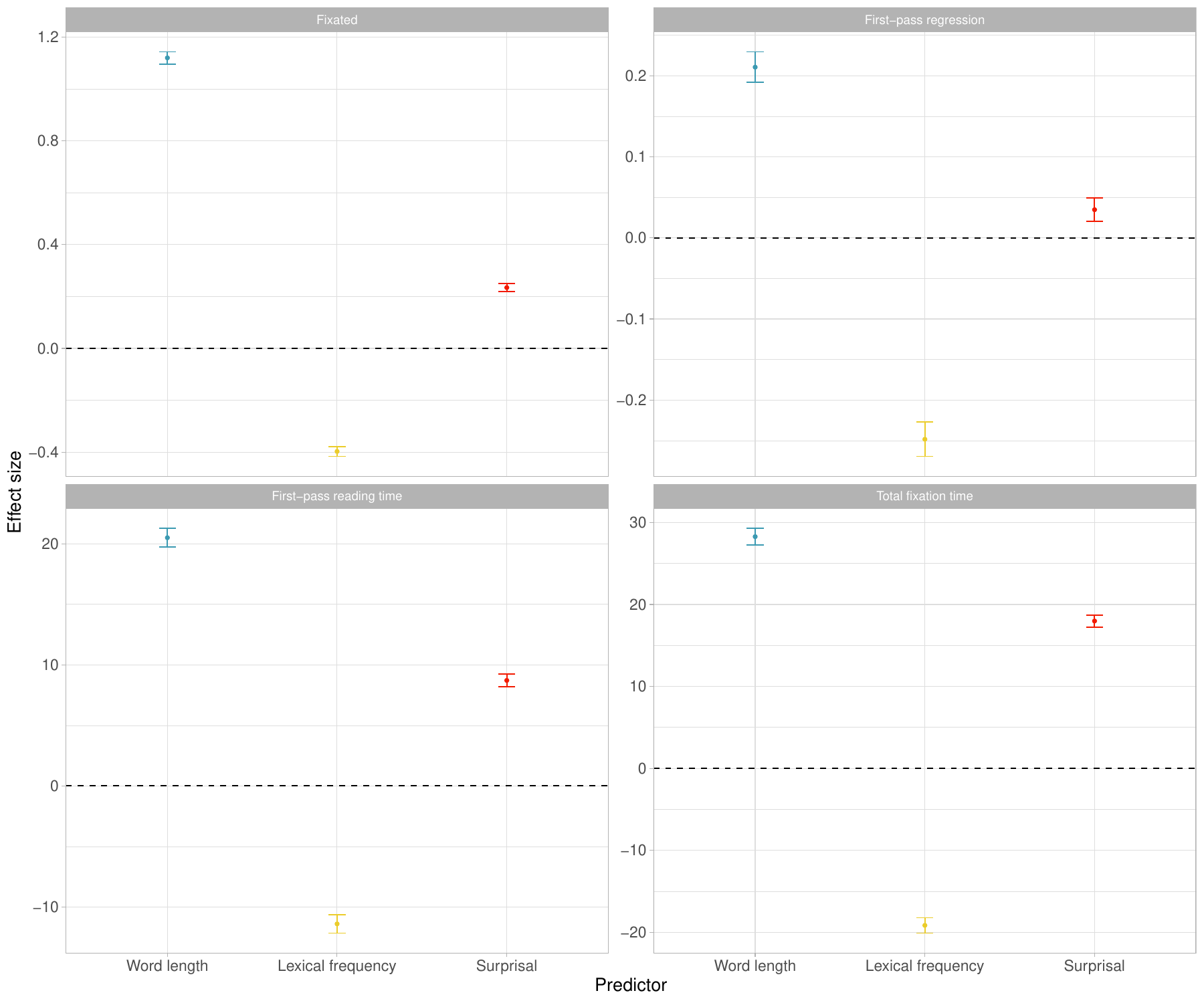}
    \caption{Posterior distributions (mean and 95\% credible interval) over the word length, lexical frequency and surprisal coefficients, estimated via the Bayesian hierarchical model presented in Eq.~\ref{eq:lmm}. The effect sizes (y-axis) represents milliseconds for the continuous reading measures, log-odds for the binary reading measures. }
    \label{fig:pla}
\end{figure}

We present posterior distributions (mean and 95\% credible intervals) for the word length, lexical frequency and surprisal coefficients in Figure~\ref{fig:pla}. For the continuous variables, the effect sizes are in milliseconds, for the binary variables in log-odds. We see that all predictors are in line with their effect on processing effort: longer words induce longer reading times, are more likely to be fixated, and are more likely to lead to a successive regression. The same pattern can be observed for high-surprisal words as well as for low-frequency words.

\section{Accessing the Data}
\label{sec:accessing-the-data}

Since EMTeC consists of different types of data, we have opted for distributing it across different channels. All code implementations are available via this \href{https://github.com/DiLi-Lab/EMTeC/}{GitHub repository}\footnote{https://github.com/DiLi-Lab/EMTeC/}. This repository can be cloned and the eye-tracking data files can be automatically downloaded using a Python script. Alternatively, the eye-tracking data can also be downloaded via this \href{https://osf.io/ajqze/}{Open Science Framework (OSF) repository}\footnote{https://osf.io/ajqze/}. The transition scores, attention scores, and hidden states are very large tensors and require a a large amount of disk space to be stored.\footnote{The transition scores amount to around 32\,GB; the hidden states to 114\,GB; and the attention scores to 193\,GB.} We make these available via this \href{https://dataverse.harvard.edu/dataset.xhtml?persistentId=doi:10.7910/DVN/GCU0W8}{Harvard Dataverse Dataset}\footnote{https://dataverse.harvard.edu/dataset.xhtml?persistentId=doi:10.7910/DVN/GCU0W8}.

\section{Summary \& Conclusion}
\label{sec:conclusion}

We presented EMTeC, the \textbf{E}ye Movements on \textbf{M}achine-Generated \textbf{Te}xts \textbf{C}orpus, a naturalistic English eye-tracking-while-reading corpus. EMTeC is the first eye-tracking corpus whose experimental stimuli were generated with different large language models utilizing five different decoding strategies. The language models are of different size and belong to different model families, and the texts can be categorized into six different types, including fiction, poetry, and summarization. The corpus further entails the language model internals during generation, \emph{i.e.}, the transition scores, the attention scores, and the hidden states. Moreover, the stimuli have been linguistically annotated with a variety of features, both on text- as well as on word-level. The eye movement data is made available at all stages of pre-processing, that is, the raw coordinate data, the fixation sequences, and the reading measures. To establish transparency and ensure reproducibility, we also release the code accompanying the corpus, \emph{i.e.}, the scripts used for the generation of the stimuli, for the linguistic annotation, for the pre-processing of the eye movement data, and for the psycholinguistic analysis. We further also provide both the uncorrected and a corrected version of the fixation sequences and the reading measures computed from them. This --- \emph{i.e.}, the data at all pre-processing stages, the code, and the corrected and uncorrected data version --- provides the user with maximal flexibility when choosing the data type that best suits their purposes and research questions, as well as allowing for the application of pre-processing algorithms different from the ones we have employed.

Based on the features described above and the nature of the corpus, we anticipate EMTeC to be utilized for a variety of use cases that range from the development of pre-processing and gaze event detection algorithms to research on vision and oculomotor control, the training and evaluation of generative models of eye movements in reading, the development of scanpath similarity metrics, the cognitive enhancement and the cognitive interpretability of language models, psycholinguistic analyses of the scanpaths and the reading measures as well as the development and evaluation of psycholinguistic theories, the investigation of the predictive power of surprisal, the investigation of text difficulty, the training and evaluation of drift correction models, and research on reading behavior as a function of the type of text that is being read.  

This list of use cases is non-exhaustive, and we envision EMTeC to be used for more intents and purposes that go beyond NLP, psycholinguistics, and reading research. EMTeC falls into a canon of a new standard of publishing eye-tracking data at multiple stages of its development and reporting its quality, and it aims to further the collection and publication of eye-tracking data in a similar manner, complementing existing data.

\paragraph{Acknowledgements and Funding}
This work was partially funded by the Swiss National Science Foundation (SNSF) under grant IZCOZ0\_220330/1 (EyeNLG), and by the German Federal Ministry of Education and Research under grant 01$\vert$ S20043 (AEye).

\paragraph{Open Practices Statement}
Materials and code are available in a GitHub repository at \url{https://github.com/DiLi-Lab/EMTeC/}. The eye-tracking data files are stored in an OSF repository at \url{https://osf.io/ajqze/}, but can be downloaded via the GitHub repository. The tensors are stored in a Harvard Dataverse at \url{https://dataverse.harvard.edu/dataset.xhtml?persistentId=doi:10.7910/DVN/GCU0W8} and can also be downloaded via the GitHub repository. None of the reported studies were preregistered.

\clearpage

\backmatter

\section*{Declarations}

\subsection*{Funding}
This work was partially funded by the Swiss National Science Foundation (SNSF) under grant IZCOZ0\_220330/1 (EyeNLG), and by the German Federal Ministry of Education and Research under grant 01$\vert$ S20043 (AEye).

\subsection*{Conflict of interest/Competing interests}
\textit{Not applicable.}

\subsection*{Ethics approval and consent to participate}
Participants in the eye-tracking experiment signed a form of informed consent, agreeing to participate in the experiment. Participation was voluntary and participants were allowed to opt out of the experiment at any time.

\subsection*{Consent for publication}
Participants in the eye-tracking experiment signed a form of informed consent, agreeing for their data to be published in anonymous form.

\subsection*{Data availability}
The data is available via this \href{https://github.com/DiLi-Lab/EMTeC/}{GitHub repository}\footnote{\url{https://github.com/DiLi-Lab/EMTeC/}}. Alternatively, the eye-tracking data can be downloaded via this \href{https://osf.io/ajqze/}{Open Science Framework (OSF) repository}\footnote{\url{https://osf.io/ajqze/}}, and the model internals via this \href{https://dataverse.harvard.edu/dataset.xhtml?persistentId=doi:10.7910/DVN/GCU0W8}{Harvard Dataverse repository}\footnote{\url{https://dataverse.harvard.edu/dataset.xhtml?persistentId=doi:10.7910/DVN/GCU0W8}}.

\subsection*{Materials availability}
The stimulus materials can be accessed via the GitHub repository \url{https://github.com/DiLi-Lab/EMTeC/} or alternatively via the Open Science Framework (OSF): \url{https://osf.io/ajqze/}. 

\subsection*{Code availability}
The code to reproduce the stimulus generation, data pre-processing, and analyses is available via the GitHub repository \url{https://github.com/DiLi-Lab/EMTeC/}.

\begin{appendices}

\section{Stimuli}
\label{appendix:stimuli}

\subsection{Model Choice}
\label{appendix:model-choice}

We opted for Phi-2~\citep{javaheripi2023phi}, Mistral 7B Instruct~\citep{jiang2023mistral} and WizardLM~\citep{xu2023wizardlm} as they yielded the best output from an eye-tracking experiment point-of-view. Since the participants were not informed about the fact that all texts were machine-generated, exclusion criteria entailed outputs which were frequently self-referential (e.g., \textit{As an AI model, ...}), repetition of the prompt in the output, and repetitions within the output. We not only wanted to have eye movements during natural reading, which would be disturbed by sentences being repeated, but we also did not want participants to know about the nature of the texts, lest they might be primed to search for AI cues.

We experimented with other models: GPT-2~\citep{radford2019language} (the deterministic outputs, \emph{i.e.}, greedy search and beam search, were not suitable as they contained a lot of repetition; Falcon-40B Instruct~\citep{almazrouei2023falcon} (non-sensical and self-referential output); Llama-2-70B Instruct~\citep{touvron2023llama2} (computationally not feasibly); XGen~\citep{nijkamp2023xgen} (the output would have been suitable from a quality perspective, but the model did not predict white-space characters, so truncating and pooling to word level would have been difficult); and MPT-30B-Instruct~\citep{mpt}(irregularities in the prediction of white-space characters that made truncation and pooling difficult).

\subsection{Prompts}
\label{appendix:prompts}

\begin{figure}[htb!]
  \centering
  \includegraphics[width=0.6\linewidth]{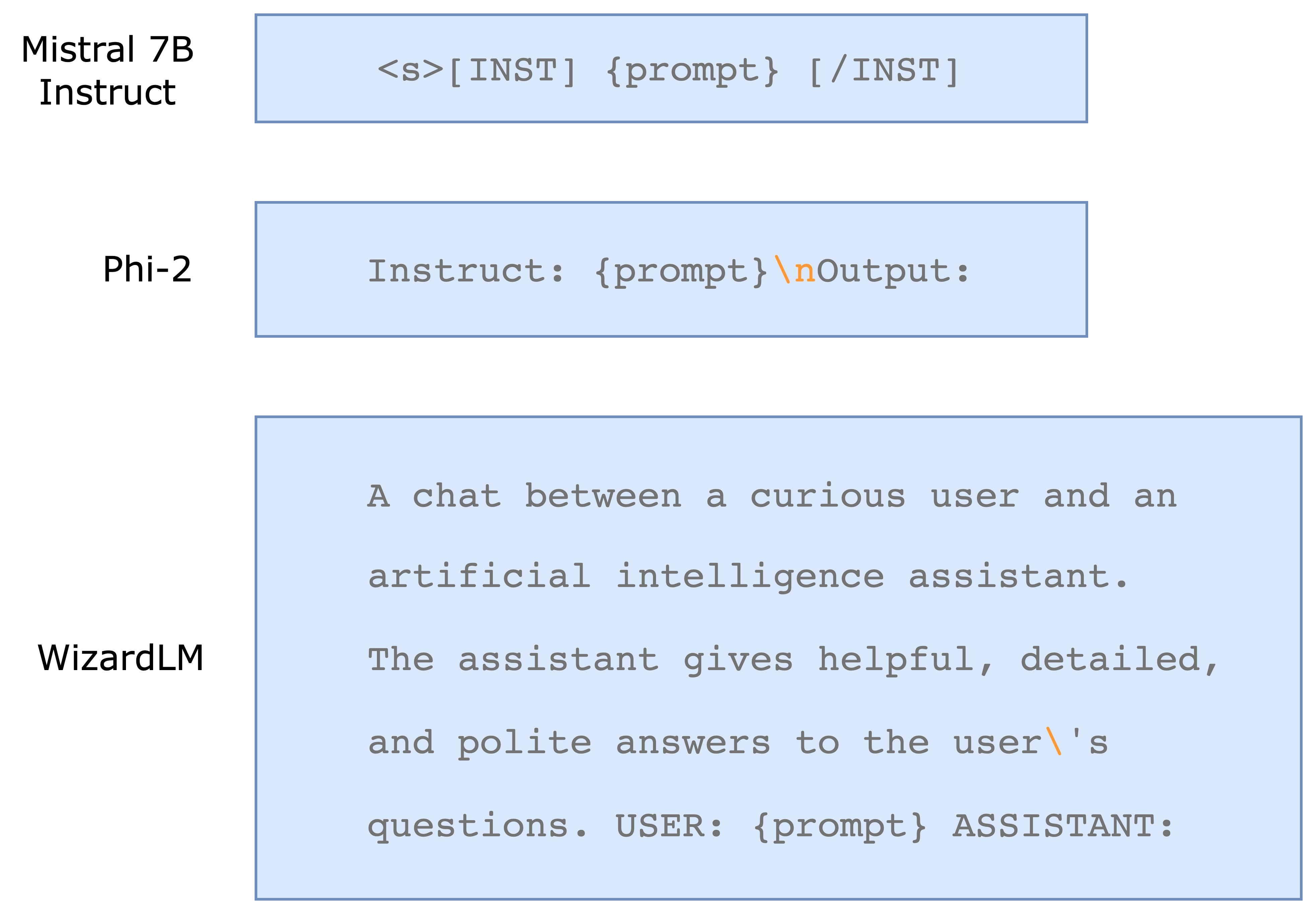}
  
  \caption{Prompt templates used for Mistral 7B Instruct, Phi-2, and WizardLM. The curly brackets \{prompt\} indicate the placeholder for the actual prompt.}
  \label{fig:prompt-templates}
\end{figure}

\subsection{Comprehension Question Generation}
\label{appendix:comprehension-question-generation}

\begin{figure}[htbp]
    \centering
    \includegraphics[width=0.95\linewidth]{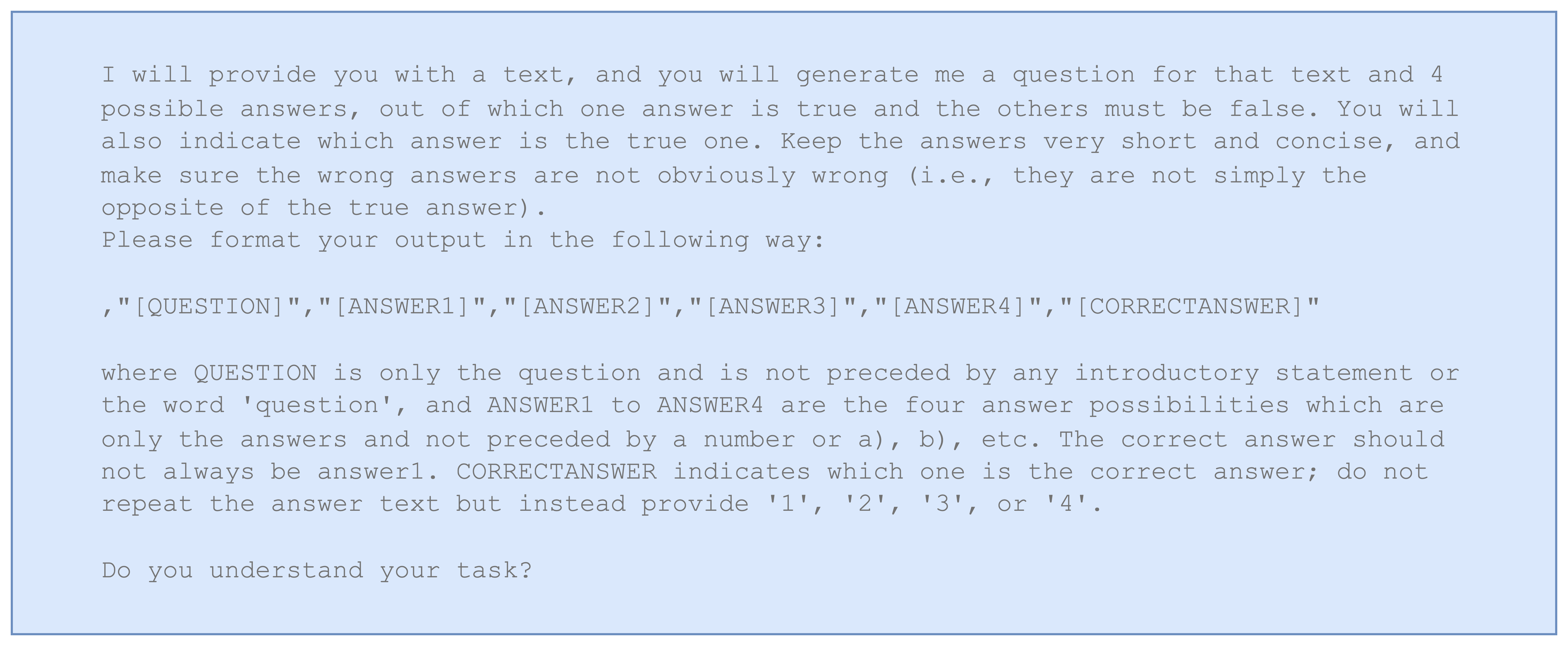}
    \caption{Prompt template used for ChatGPT to generate the comprehension questions.}
    \label{fig:chatgpt-template}
\end{figure}

\subsection{Descriptive Statistics of the Stimuli}
\label{appendix:descriptive-stats-stimuli}

\begin{table}[h]
\caption{Descriptive statistics of the average text length, word length, Zipf frequency, Flesch reading ease, and GPT-2 surprisal for each decoding strategy used by Mistral.}
\label{tab:stats-text-mistral-all-dec}
\begin{tabular}{lllllll}
\hline
                                &      & Text Length & Word Length & Zipf Freq. & Flesch  & Surp. GPT-2 \\ \hline
\multirow{4}{*}{Greedy search} & mean & 89.881      & 5.11        & 5.678      & 49.261  & 3.891       \\
                                & std  & 19.042      & 0.509       & 0.247      & 19.32   & 3.653       \\
                                & min  & 44.0        & 3.621       & 5.153      & 17.836  & 0.0         \\
                                & max  & 118.0       & 5.899       & 6.406      & 99.814  & 30.897      \\ \hline
\multirow{4}{*}{Beam search}   & mean & 90.548      & 5.137       & 5.641      & 47.689  & 3.91        \\
                                & std  & 19.267      & 0.473       & 0.237      & 18.688  & 3.748       \\
                                & min  & 44.0        & 4.086       & 4.943      & 16.493  & 0.0         \\
                                & max  & 114.0       & 5.947       & 6.045      & 100.731 & 30.897      \\ \hline
\multirow{4}{*}{Sampling}       & mean & 88.595      & 5.176       & 5.634      & 46.114  & 4.031       \\
                                & std  & 18.456      & 0.461       & 0.222      & 17.547  & 3.786       \\
                                & min  & 37.0        & 4.362       & 4.999      & 5.949   & 0.0         \\
                                & max  & 117.0       & 6.66        & 6.112      & 83.854  & 30.897      \\ \hline
\multirow{4}{*}{Top-$k$}           & mean & 92.0        & 5.16        & 5.638      & 47.107  & 3.963       \\
                                & std  & 19.44       & 0.432       & 0.235      & 17.733  & 3.707       \\
                                & min  & 37.0        & 4.362       & 4.999      & 6.508   & 0.0         \\
                                & max  & 124.0       & 5.949       & 6.112      & 83.854  & 30.897      \\ \hline
\multirow{4}{*}{Top-$p$}           & mean & 88.952      & 5.192       & 5.622      & 43.891  & 3.881       \\
                                & std  & 20.889      & 0.447       & 0.236      & 17.556  & 3.671       \\
                                & min  & 37.0        & 4.141       & 5.146      & 6.194   & 0.0         \\
                                & max  & 119.0       & 6.33        & 6.191      & 89.08   & 30.897      \\ \hline
\end{tabular}
\end{table}

\begin{table}[h]
\caption{Descriptive statistics of the average text length, word length, Zipf frequency, Flesch reading ease, and GPT-2 surprisal for each decoding strategy used by Phi-2.}
\label{tab:stats-text-phi2-all-dec}
\begin{tabular}{llllllll}
\hline
                                &      & Text Length & Word Length & Zipf Freq. & Flesch  & Surp. GPT-2 \\ \hline
\multirow{4}{*}{Greedy search} & mean & 82.595      & 5.101       & 5.726      & 47.348  & 3.813       \\
                                & std  & 22.568      & 0.547       & 0.18       & 21.134  & 3.64        \\
                                & min  & 41.0        & 3.901       & 5.366      & 11.623  & 0.0         \\
                                & max  & 126.0       & 6.032       & 6.189      & 103.438 & 30.897      \\ \hline
\multirow{4}{*}{Sampling}       & mean & 83.024      & 5.211       & 5.648      & 44.002  & 4.1         \\
                                & std  & 27.542      & 0.603       & 0.251      & 22.009  & 3.8         \\
                                & min  & 34.0        & 3.853       & 4.889      & 16.549  & 0.0         \\
                                & max  & 127.0       & 6.411       & 6.094      & 105.304 & 30.897      \\ \hline
\multirow{4}{*}{Top-$k$}           & mean & 85.714      & 5.114       & 5.687      & 47.396  & 4.052       \\
                                & std  & 28.905      & 0.59        & 0.213      & 22.216  & 3.759       \\
                                & min  & 30.0        & 3.804       & 5.128      & 18.189  & 0.0         \\
                                & max  & 128.0       & 6.411       & 6.083      & 105.304 & 30.897      \\ \hline
\multirow{4}{*}{Top-$p$}           & mean & 85.976      & 5.098       & 5.684      & 46.856  & 4.035       \\
                                & std  & 26.321      & 0.558       & 0.217      & 21.865  & 3.703       \\
                                & min  & 25.0        & 3.795       & 4.952      & 14.999  & 0.0         \\
                                & max  & 130.0       & 6.028       & 6.086      & 105.246 & 29.314      \\ \hline
\end{tabular}
\end{table}

\begin{table}[h]
\caption{Descriptive statistics of the average text length, word length, Zipf frequency, Flesch reading ease, and GPT-2 surprisal for each decoding strategy used by WizardLM.}
\label{tab:stats-text-wizardlm-all-dec}
\begin{tabular}{lllllll}
\hline
                               &      & Text Length & Word Length & Zipf Freq. & Flesch  & Surp. GPT-2 \\ \hline
\multirow{4}{*}{Greedy search} & mean & 84.833      & 5.118       & 5.623      & 46.237  & 4.037       \\
                               & std  & 18.939      & 0.535       & 0.246      & 20.407  & 3.798       \\
                               & min  & 40.0        & 4.184       & 4.951      & 7.097   & 0.0         \\
                               & max  & 118.0       & 6.012       & 6.041      & 94.045  & 30.897      \\ \hline
\multirow{4}{*}{Beam search}   & mean & 82.881      & 5.167       & 5.618      & 44.762  & 3.984       \\
                               & std  & 14.737      & 0.552       & 0.225      & 22.727  & 3.812       \\
                               & min  & 49.0        & 3.903       & 4.827      & 8.631   & 0.0         \\
                               & max  & 118.0       & 6.012       & 6.035      & 103.673 & 24.775      \\ \hline
\multirow{4}{*}{Sampling}      & mean & 85.833      & 5.239       & 5.606      & 44.131  & 4.22        \\
                               & std  & 14.385      & 0.508       & 0.228      & 20.123  & 3.919       \\
                               & min  & 48.0        & 4.077       & 5.015      & 9.981   & 0.0         \\
                               & max  & 111.0       & 6.303       & 6.117      & 108.073 & 29.314      \\ \hline
\multirow{4}{*}{Top-$k$}       & mean & 85.19       & 5.221       & 5.598      & 45.118  & 4.262       \\
                               & std  & 17.444      & 0.542       & 0.248      & 20.753  & 3.95        \\
                               & min  & 28.0        & 4.015       & 5.103      & 10.783  & 0.0         \\
                               & max  & 114.0       & 6.405       & 5.979      & 108.073 & 30.897      \\ \hline
\multirow{4}{*}{Top-$p$}       & mean & 85.19       & 5.199       & 5.613      & 45.188  & 4.097       \\
                               & std  & 15.121      & 0.529       & 0.212      & 21.432  & 3.806       \\
                               & min  & 44.0        & 4.156       & 5.08       & 4.783   & 0.0         \\
                               & max  & 112.0       & 6.277       & 5.986      & 104.338 & 30.897      \\ \hline
\end{tabular}
\end{table}

\FloatBarrier

\section{Priors for Bayesian hierarchical generalized linear-mixed models}
\label{appendix:priors}

\begin{table}[h!]
\centering
\begin{tabular}{l l}
\toprule
\bf Log-linear regression model & \bf Logistic regression model \\
\midrule
$\beta_0\sim \mathcal{N}(6, 1.5)$ & $\beta_0\sim \mathcal{N}(0, 4)$ \\
$\beta_1\dots\beta_4 \sim \mathcal{N}(0,1)$ & $\beta_1\dots\beta_4 \sim \mathcal{N}(0,1)$ \\
$\sigma\sim \mathcal{N}(0, 1)$ & \\
\bottomrule
\end{tabular}
\caption[Prior distributions]{Overview of model priors.}
\end{table}

\section{Manual Fixation Correction}
\label{appendix:manual-fixation-correction}

\begin{figure}[h]
    \centering
    \includegraphics[width=0.9\textwidth]{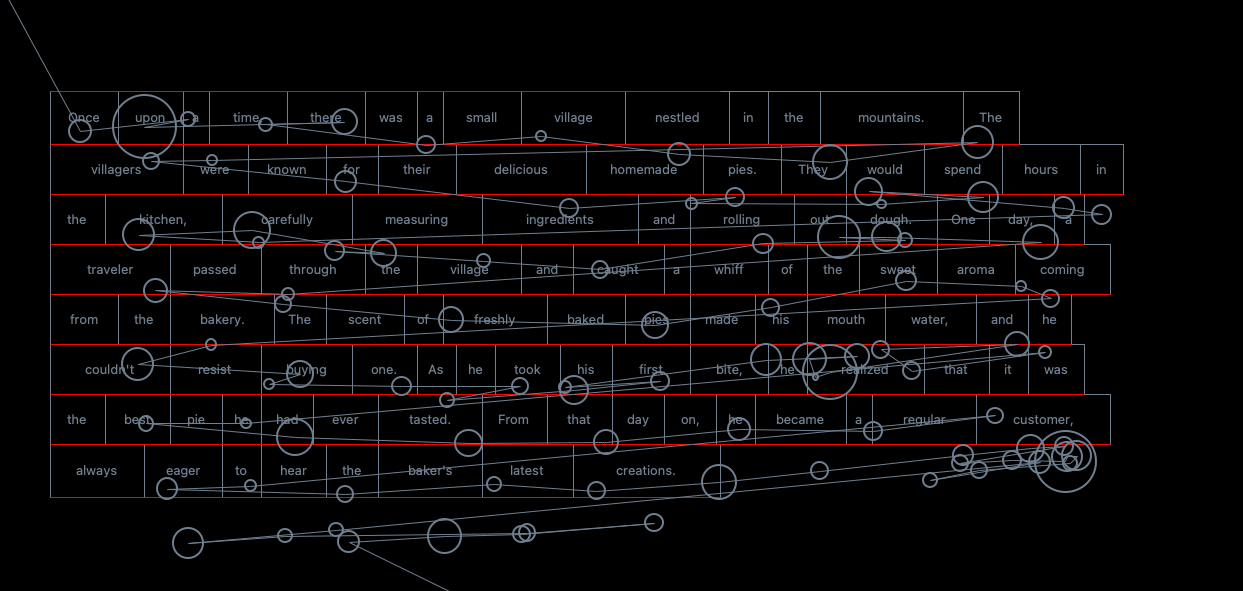}
    \caption{An example of vertical drift during the eye-tracking experiment. Subject 28 reading item06.}
    \label{fig:vertical-drift-et28-item06}
\end{figure}

\begin{figure}[h]
    \centering
    \includegraphics[width=0.9\textwidth]{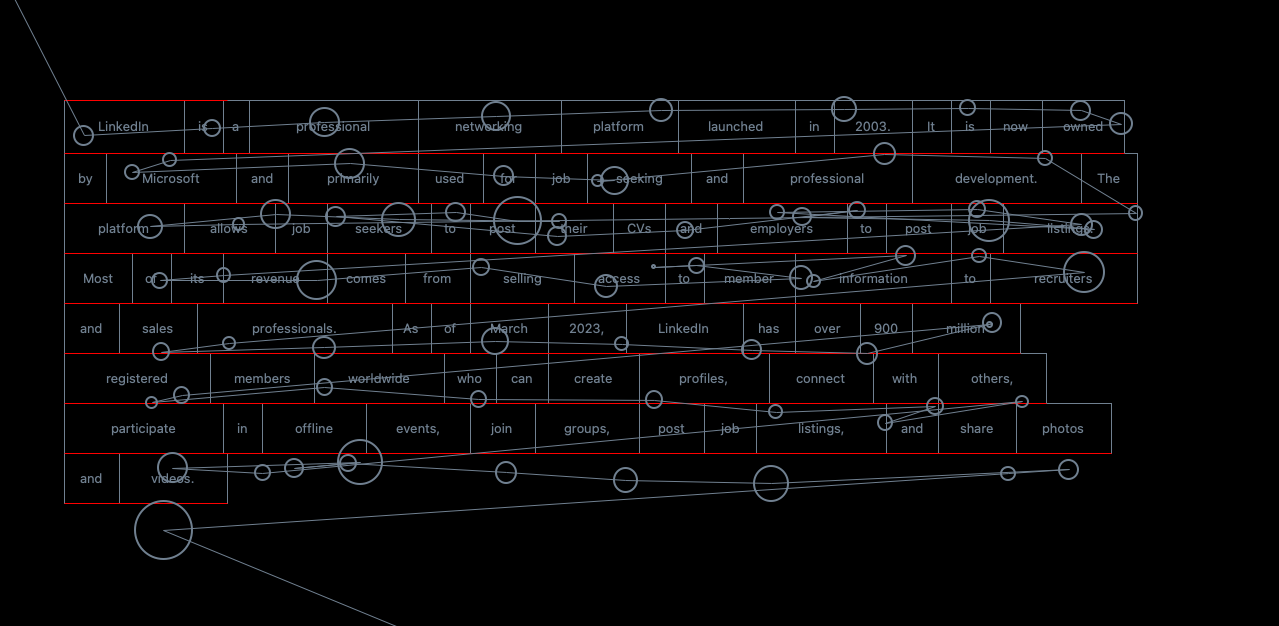}
    \caption{An example of vertical drift during the eye-tracking experiment. Subject 28 reading item35.}
    \label{fig:vertical-drift-et28-item35}
\end{figure}

\begin{figure}[h]
    \centering
    \includegraphics[width=0.9\textwidth]{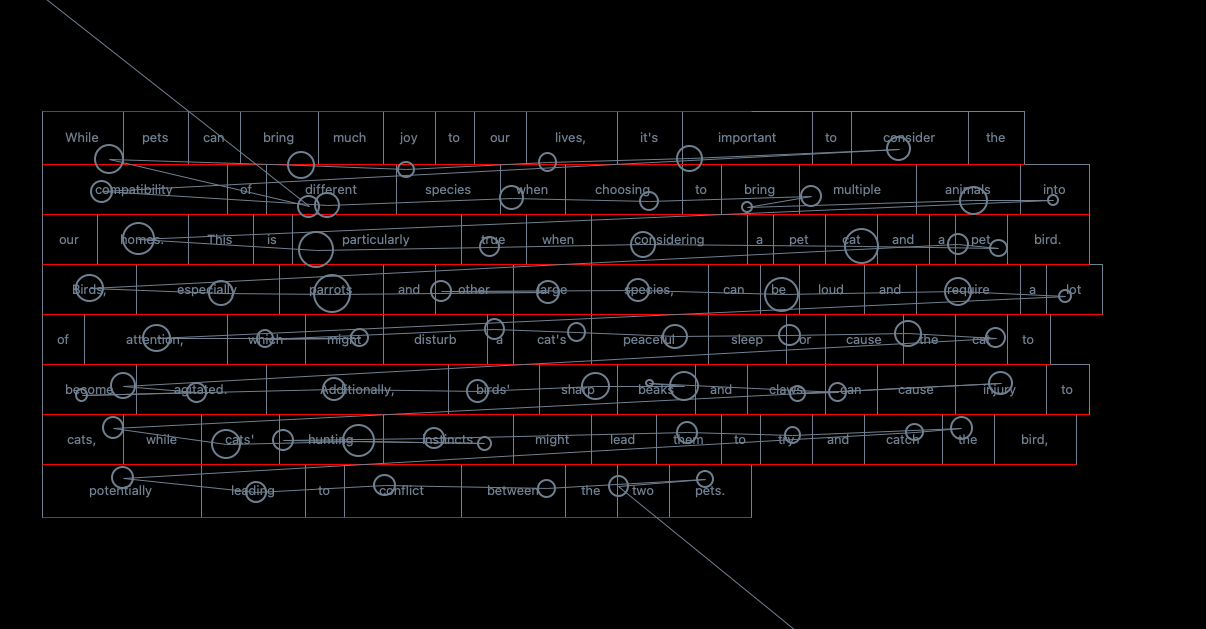}
    \caption{An example of vertical drift during the eye-tracking experiment. Subject 42 reading item01.}
    \label{fig:vertical-drift-et42-item01}
\end{figure}

\begin{figure}[h]
    \centering
    \includegraphics[width=0.9\textwidth]{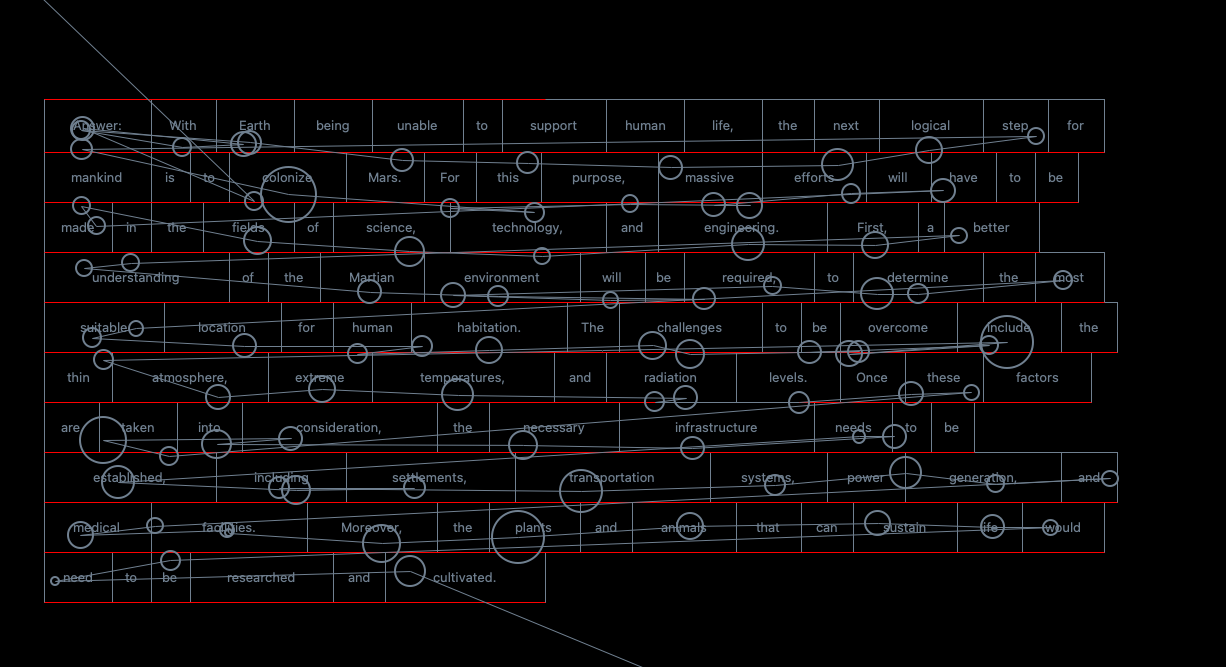}
    \caption{An example of vertical drift during the eye-tracking experiment. Subject 42 reading item72.}
    \label{fig:vertical-drift-et42-item72}
\end{figure}

\begin{figure}[h]
    \centering
    \includegraphics[width=0.9\textwidth]{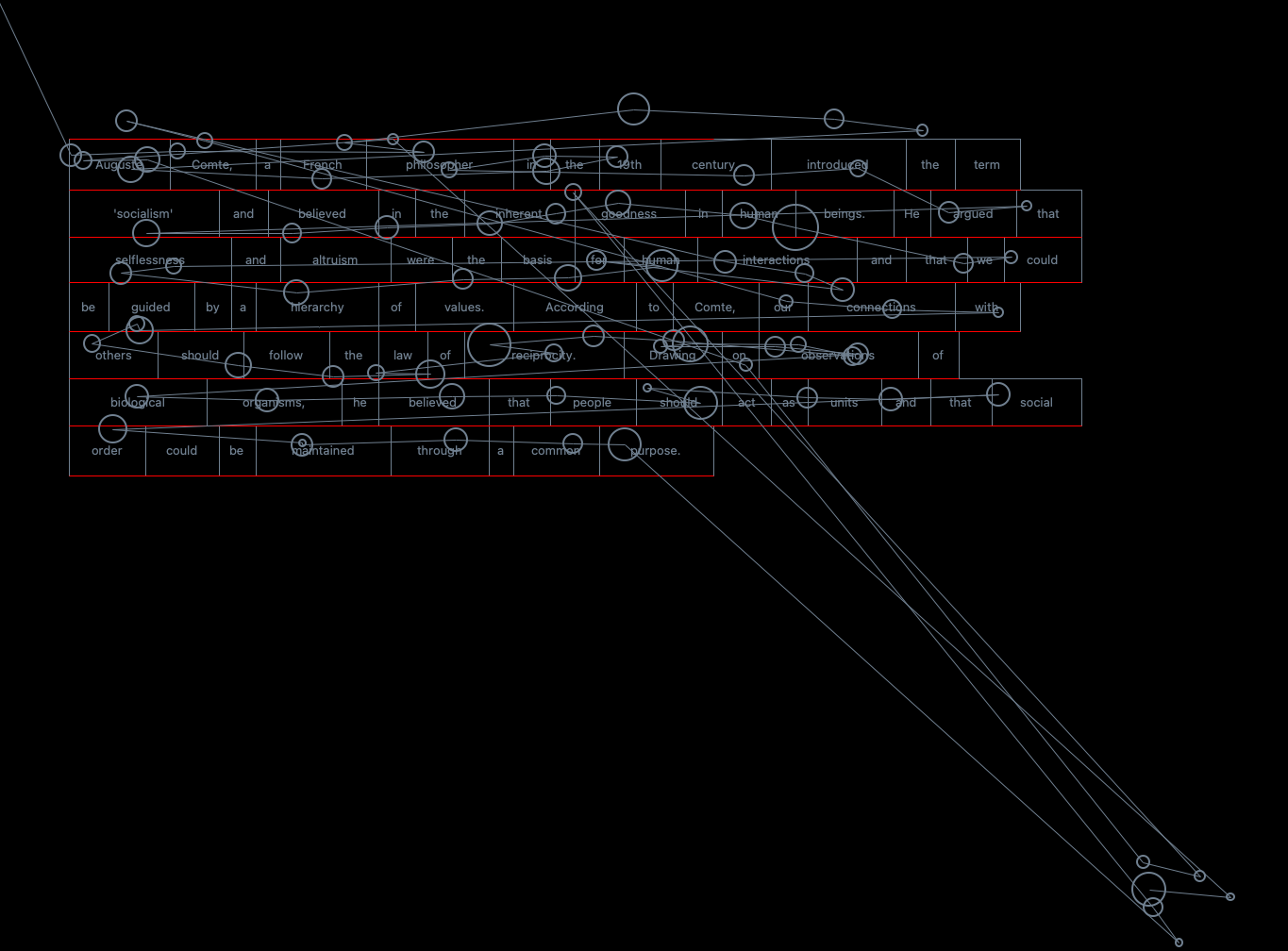}
    \caption{An example of vertical drift during the eye-tracking experiment, including jumps out of the text to the blue sticker and back into the text. Subject 49 reading item51.}
    \label{fig:vertical-drift-et49-item51}
\end{figure}

\begin{figure}[h]
    \centering
    \includegraphics[width=0.9\textwidth]{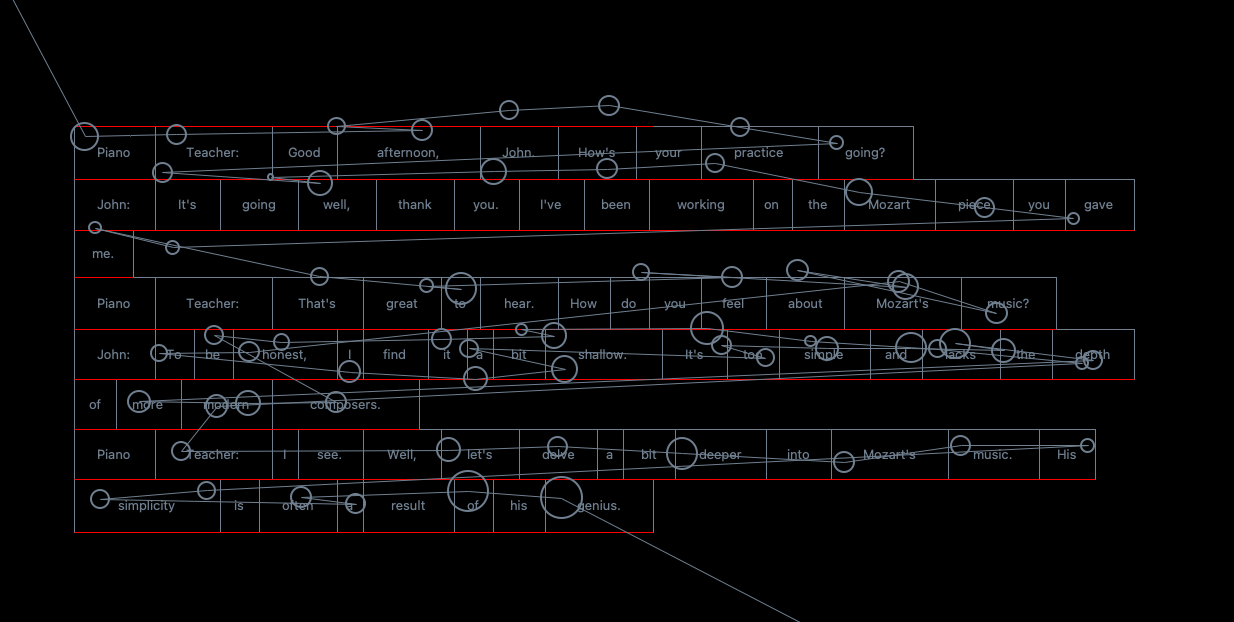}
    \caption{An example of vertical drift during the eye-tracking experiment. Subject 49 reading item80.}
    \label{fig:vertical-drift-et49-item80}
\end{figure}

\begin{figure}[h]
    \centering
    \includegraphics[width=0.9\textwidth]{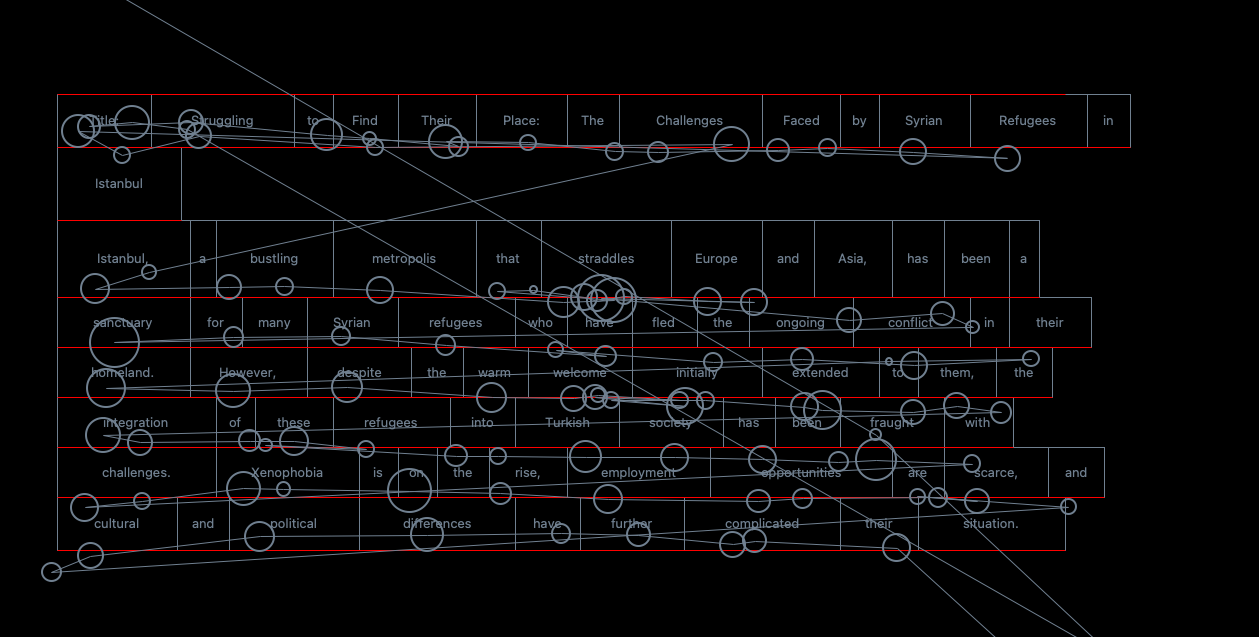}
    \caption{An example of vertical drift during the eye-tracking experiment. Subject 50 reading item46.}
    \label{fig:vertical-drift-et50-item46}
\end{figure}

\begin{figure}[h]
    \centering
    \includegraphics[width=0.9\textwidth]{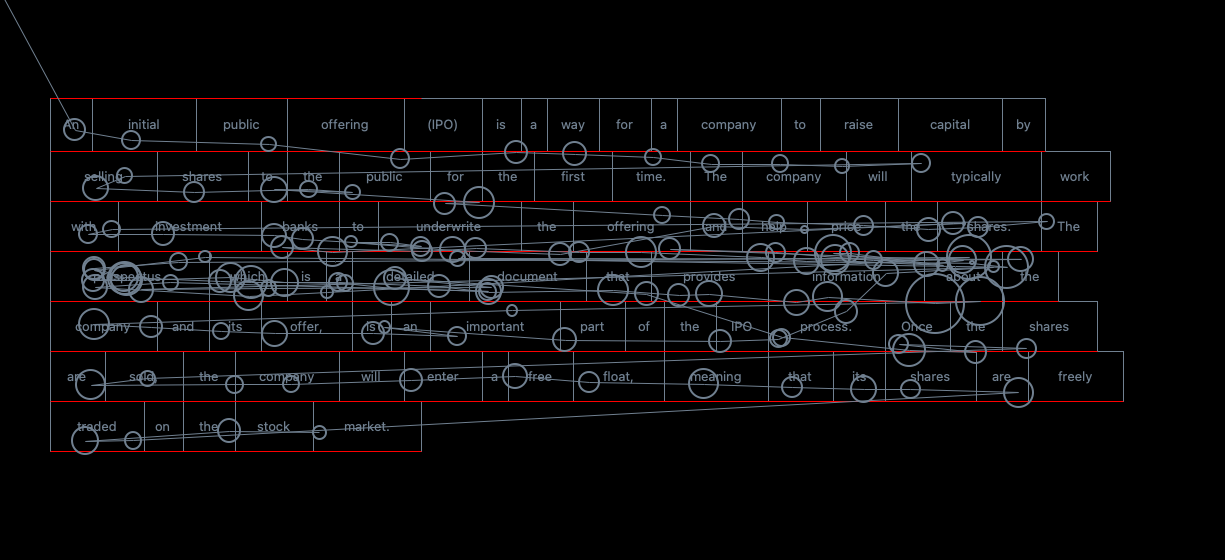}
    \caption{An example of vertical drift during the eye-tracking experiment. Subject 50 reading item53.}
    \label{fig:vertical-drift-et50-item53}
\end{figure}

\begin{figure}[h]
    \centering
    \includegraphics[width=0.9\textwidth]{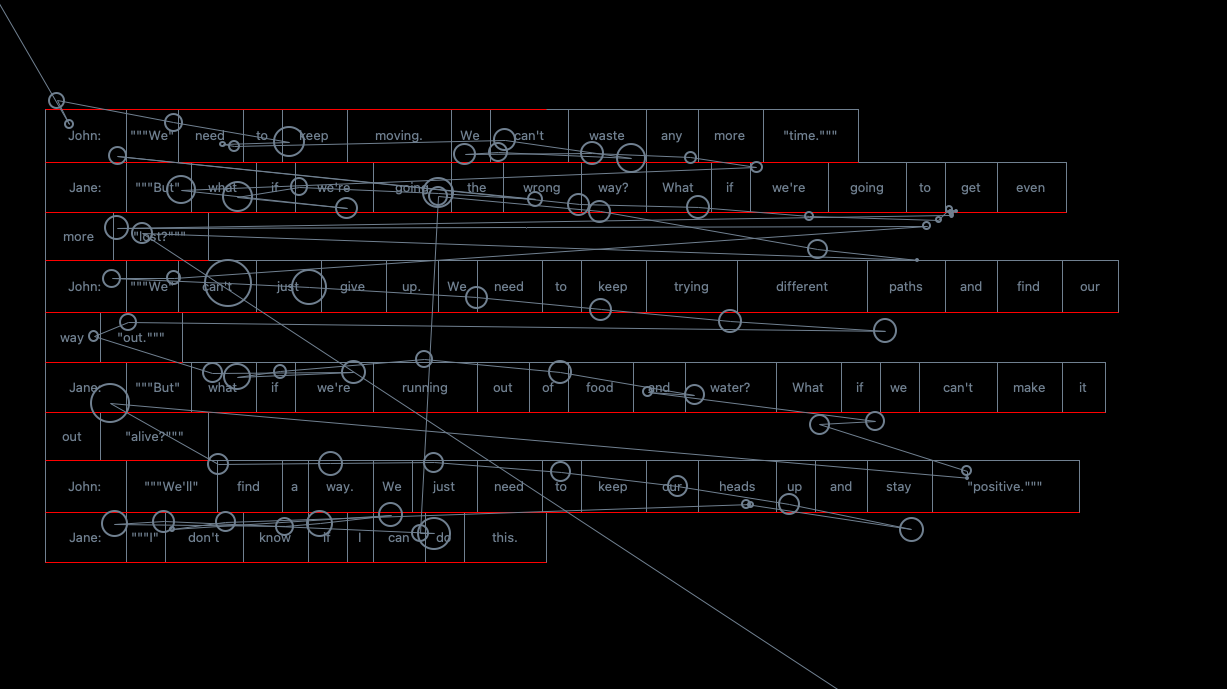}
    \caption{An example of vertical drift during the eye-tracking experiment. Subject 67 reading item11.}
    \label{fig:vertical-drift-et67-item11}
\end{figure}

\begin{figure}[h]
    \centering
    \includegraphics[width=0.9\textwidth]{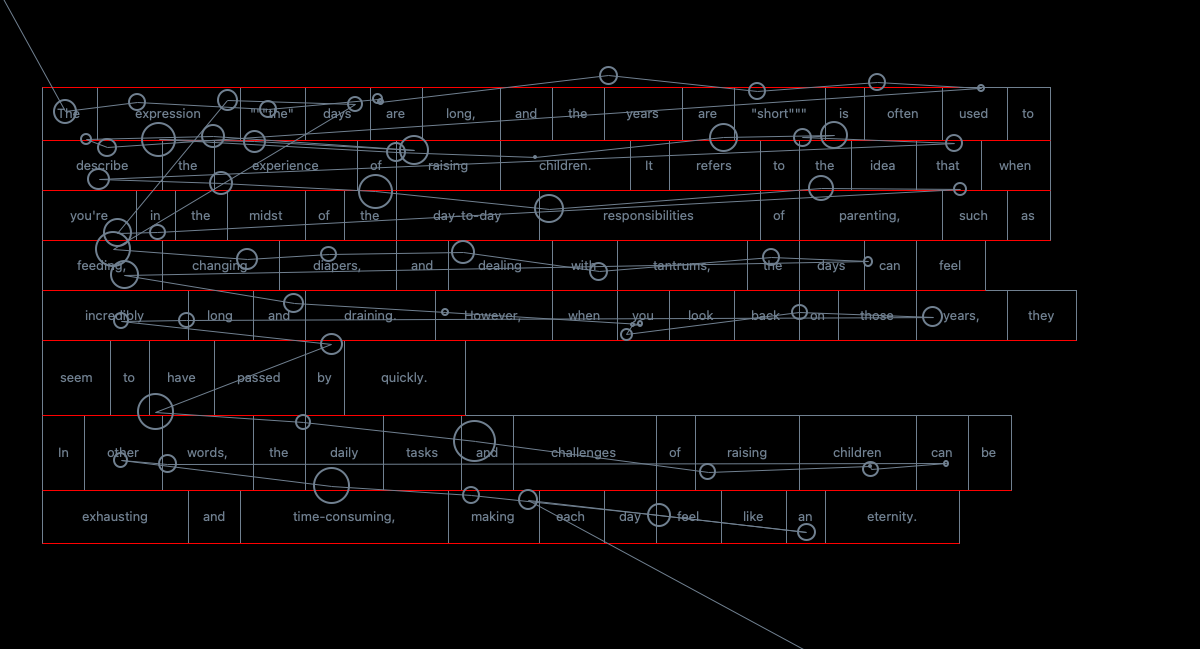}
    \caption{An example of vertical drift during the eye-tracking experiment. Subject 67 reading item14.}
    \label{fig:vertical-drift-et67-item14}
\end{figure}

\begin{figure}[h]
    \centering
    \includegraphics[width=0.9\textwidth]{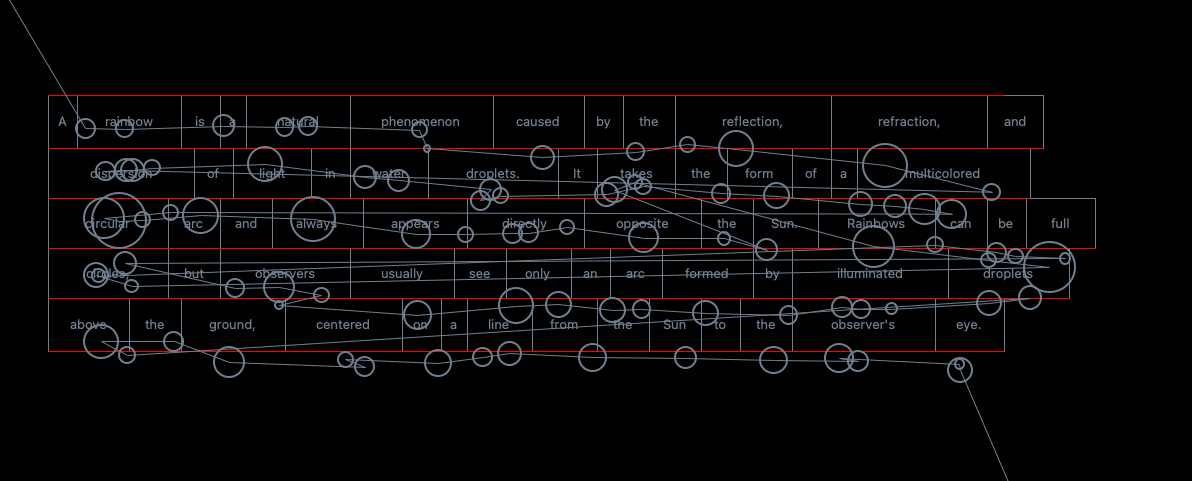}
    \caption{An example of vertical drift during the eye-tracking experiment. Subject 81 reading item40.}
    \label{fig:vertical-drift-et81-item40}
\end{figure}

\begin{figure}[h]
    \centering
    \includegraphics[width=0.9\textwidth]{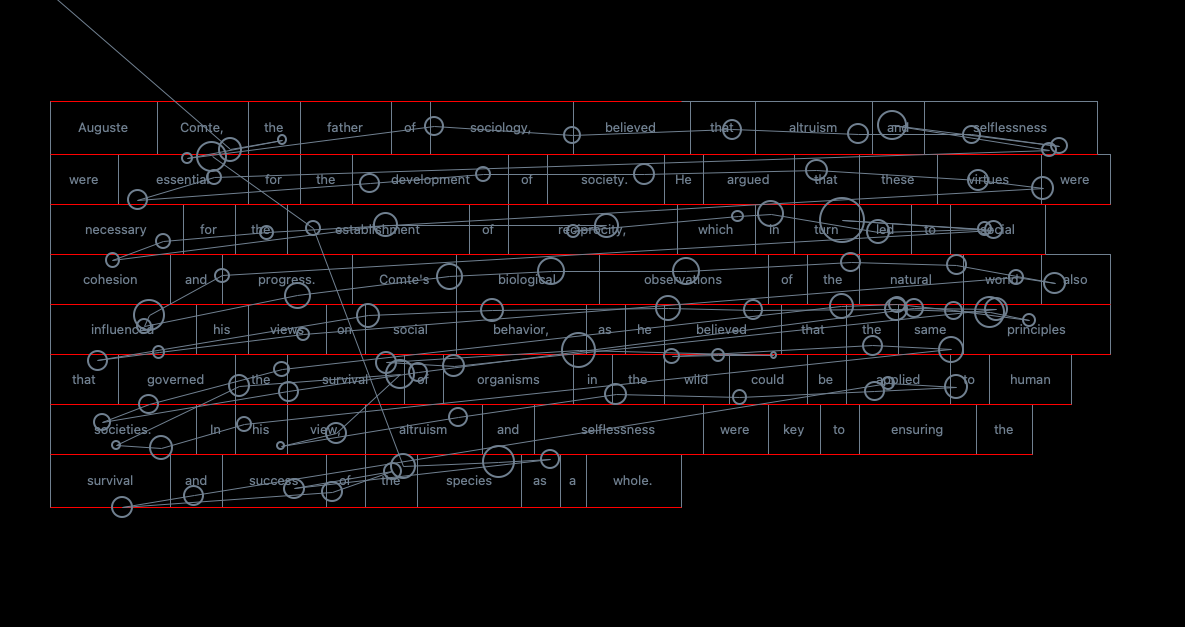}
    \caption{An example of vertical drift during the eye-tracking experiment. Subject 83 reading item51.}
    \label{fig:vertical-drift-et83-item51}
\end{figure}




\end{appendices}


\clearpage

\bibliography{emtec}

\end{document}